\documentclass{article}

\usepackage{arxiv}

\usepackage[utf8]{inputenc} 
\usepackage[T1]{fontenc}    
\usepackage{url}            
\usepackage{booktabs}       

\usepackage{amsmath,amssymb,amsfonts,latexsym,amsthm,bm,breqn,dsfont,gensymb}
\usepackage{latexsym,breqn}
\usepackage{nicefrac}       
\usepackage{microtype}      
\usepackage{natbib}

\usepackage{graphicx,subfigure}

\usepackage{algorithm}
\usepackage[noend]{algpseudocode}
\makeatletter
\def\BState{\State\hskip-\ALG@thistlm}

\usepackage{ctable,multicol,multirow}

\usepackage{color}
\definecolor{green}{rgb}{0, 0.5, 0}
\usepackage[pdftex,
			bookmarks, bookmarksnumbered=true, bookmarksopen=true,
			colorlinks, citecolor=blue, filecolor=black, linkcolor=red, urlcolor=green,
			pdfauthor={López-Lopera, Andrés Felipe}, pdftitle={PhDEMSE},
			pdftoolbar=true, pdfmenubar=true, pdffitwindow = true
			]{hyperref}

\usepackage{pifont}
\definecolor{green}{rgb}{0, 0.5, 0}
\newcommand{\goodcheck}{{\color{green} \ding{51}}}
\newcommand{\badcheck}{{\color{red} \ding{55}}}

\newcommand{\Bz}{\boldsymbol{z}}

\newcommand{\Bx}{\boldsymbol{x}}
\newcommand{\By}{\boldsymbol{y}}

\newcommand{\Bzero}{\boldsymbol{0}}
\newcommand{\Beye}{\boldsymbol{I}}
\newcommand{\Bl}{\boldsymbol{l}}
\newcommand{\Bu}{\boldsymbol{u}}
\newcommand{\BGamma}{\boldsymbol{\Gamma}}
\newcommand{\BPhi}{\boldsymbol{\Phi}}
\newcommand{\BLambda}{\boldsymbol{\Lambda}}
\newcommand{\BSigma}{\boldsymbol{\Sigma}}

\newcommand{\Bvarepsilon}{\boldsymbol{\varepsilon}}
\newcommand{\Bmu}{\boldsymbol{\mu}}

\newcommand{\Bxi}{\boldsymbol{\xi}}
\newcommand{\Btheta}{\boldsymbol{\theta}}
\newcommand{\realset}[1]{\mathbb{R}^{#1}}
\newcommand{\normrnd}[2]{\mathcal{N}\left({#1,#2}\right)}
\newcommand{\tnormrnd}[4]{\mathcal{TN}\left({#1,#2,#3,#4}\right)}

\title{Approximating Gaussian Process Emulators with Linear Inequality Constraints and Noisy Observations via MC and MCMC}
\date{}
\author{
  Andr\'es F. L\'opez-Lopera\\
  Mines Saint-\'Etienne\\
  42000 Saint-\'Etienne, France \\
  \texttt{andres-felipe.lopez@emse.fr} \\
  \And
  Fran\c{c}ois Bachoc \\
  Institut de Math\'ematiques de Toulouse\\
  31062 Toulouse, France \\
  \texttt{Francois.Bachoc@math.univ-toulouse.fr} \\  
  \And
  Nicolas Durrande\\
  PROWLER.io \\
  Cambridge, CB2 1LA, UK \\
  \texttt{nicolas@prowler.io} \\
  \And
  J\'er\'emy Rohmer, D\'eborah Idier \\
  BRGM \\
  45060 Orl\'eans c\'edex 2, France \\
  \texttt{\{j.rohmer,d.idier\}@brgm.fr} \\  
  \And
  Olivier Roustant \\
  Mines Saint-\'Etienne\\
  42000 Saint-\'Etienne, France \\
  \texttt{roustant@emse.fr} \\
}

\begin{document}
\maketitle

\begin{abstract}
	Adding inequality constraints (e.g. boundedness, monotonicity, convexity) into Gaussian processes (GPs) can lead to more realistic stochastic emulators. Due to the truncated Gaussianity of the posterior, its distribution has to be approximated. In this work, we consider Monte Carlo (MC) and Markov Chain Monte Carlo (MCMC) methods. However, strictly interpolating the observations may entail expensive computations due to highly restrictive sample spaces. Furthermore, having (constrained) GP emulators when data are actually noisy is also of interest for real-world implementations. Hence, we introduce a noise term for the relaxation of the interpolation conditions, and we develop the corresponding approximation of GP emulators under linear inequality constraints. We show with various toy examples that the performance of MC and MCMC samplers improves when considering noisy observations. Finally, on 2D and 5D coastal flooding applications, we show that more flexible and realistic GP implementations can be obtained by considering noise effects and by enforcing the (linear) inequality constraints.
\end{abstract}


\section{Introduction}
\label{sec:intro}
Gaussian processes (GPs) are used in a great variety of real-world problems as stochastic emulators in fields such as biology, finance and robotics \citep{Rasmussen2005GP,Murphy2012ML}. In the latter, they can be used for emulating the dynamics of robots when experiments become expensive (e.g. time consuming or highly costly) \citep{Rasmussen2005GP}. 

Imposing inequality constraints (e.g. boundedness, monotonicity, convexity) into GP emulators can lead to more realistic profiles guided by the physics of data \citep{Golchi2015MonotoneEmulation,Maatouk2017GPineqconst,LopezLopera2017FiniteGPlinear}. Some applications where constrained GP emulators have been successfully used are computer networking (monotonicity) \citep{Golchi2015MonotoneEmulation}, econometrics (positivity or monotonicity) \citep{Cousin2016KrigingFinancial}, and nuclear safety criticality assessment (positivity and monotonicity) \citep{LopezLopera2017FiniteGPlinear}.

In \citep{Maatouk2017GPineqconst,LopezLopera2017FiniteGPlinear}, an approximation of GP emulators based on (first-order) regression splines is introduced in order to satisfy general sets of linear inequality constraints. Because of the piecewise linearity of the finite-dimensional approximation used there, the inequalities are satisfied everywhere in the input space. Furthermore, The authors of \citep{Maatouk2017GPineqconst,LopezLopera2017FiniteGPlinear} proved that the resulting posterior distribution conditioned on both observations and inequality constraints is truncated Gaussian-distributed. Finally, it was shown in \citep{Bay2016KimeldorfWahba} that the resulting posterior mode converges uniformly to the thin plate spline solution when the number of knots of the spline goes to infinity.

Since the posterior is a truncated GP, its distribution cannot be computed in closed-form, but it can be approximated via Monte Carlo (MC) or Markov chain Monte Carlo (MCMC) \citep{Maatouk2017GPineqconst,LopezLopera2017FiniteGPlinear}. Several MC and MCMC samplers have been tested in \citep{LopezLopera2017FiniteGPlinear}, leading to emulators that perform well for one or two dimensional input spaces. Starting from the claim that allowing noisy observations could yield less constrained sample spaces for samplers, here we develop the corresponding approximation of constrained GP emulators when adding noise. Moreover, (constrained) GP emulators for observations that are truly noisy are also of interest for practical implementations. We test the efficiency of various MC and MCMC samplers under 1D toy examples where models without observation noise yield impractical sampling routines. We also show that, in monotonic examples, our framework can be applied up to 5D and for thousands of observations providing high-quality effective sample sizes within reasonable running times.

This paper is organised as follows. In Section \ref{sec:contrGPs}, we introduce the finite-dimensional approximation of GP emulators with linear inequality constraints and noisy observations. In Section \ref{sec:numExperiments}, we apply our framework to synthetic examples where the consideration of noise-free observations is unworkable. We also test it on 2D and 5D coastal flooding applications. Finally, in Section \ref{sec:conclusions}, we highlight the conclusions, as well as potential future works.

\section{Gaussian Process Emulators with Linear Inequality Constraints and Noisy Observations}
\label{sec:contrGPs}
\begin{figure*}[t!]
	\centering
	\includegraphics[width=0.325\textwidth]{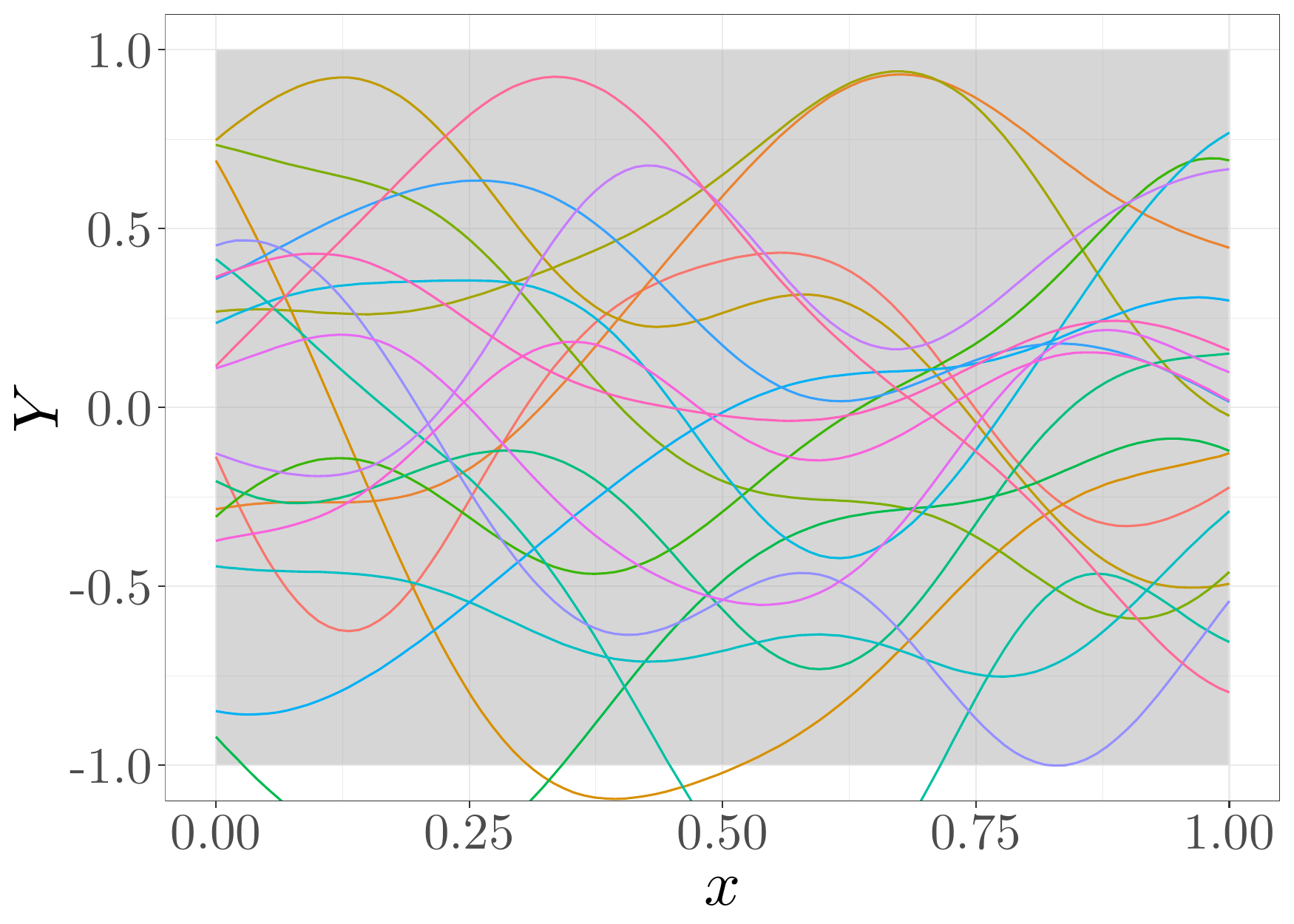}
	\includegraphics[width=0.325\textwidth]{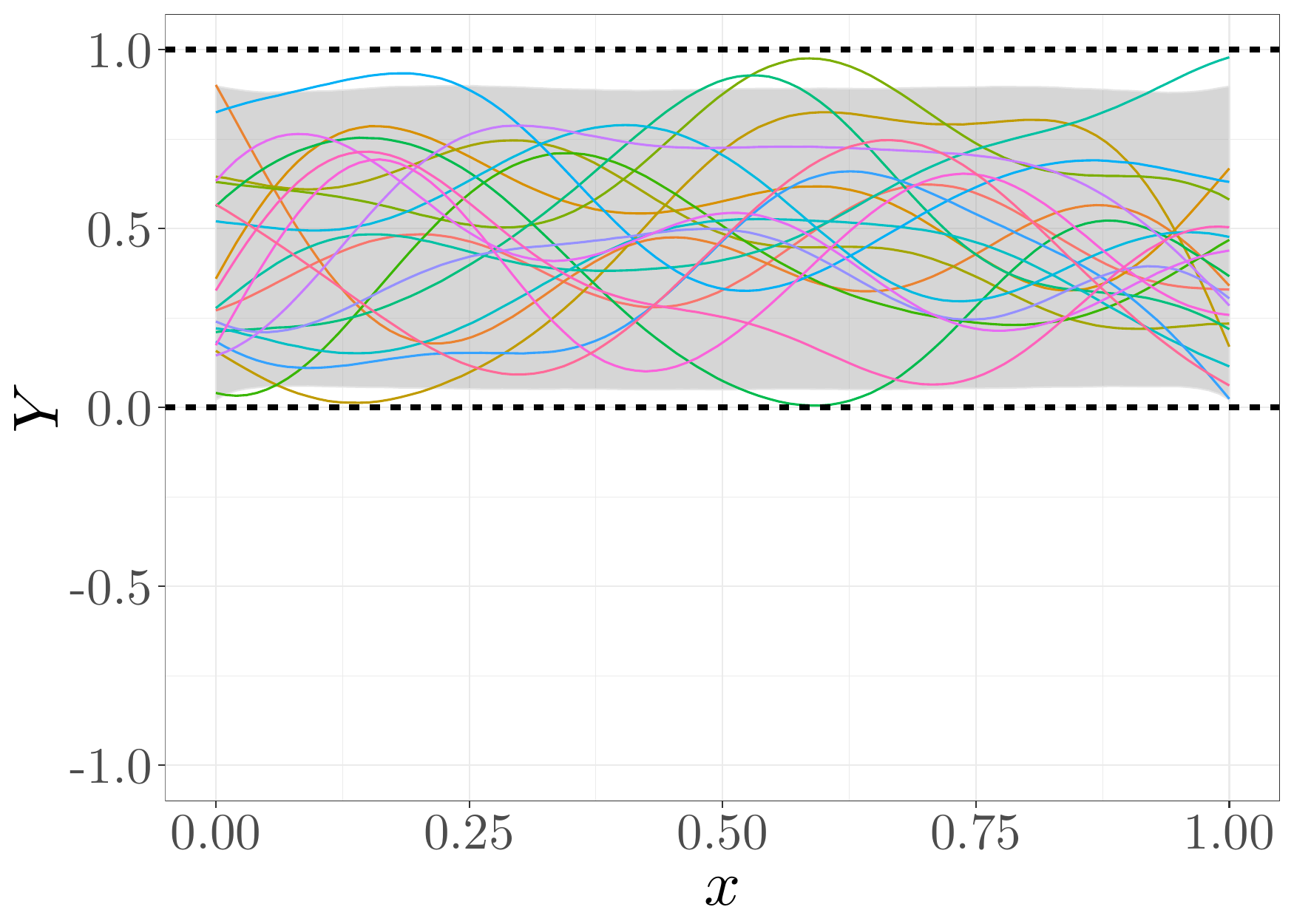}
	\includegraphics[width=0.325\textwidth]{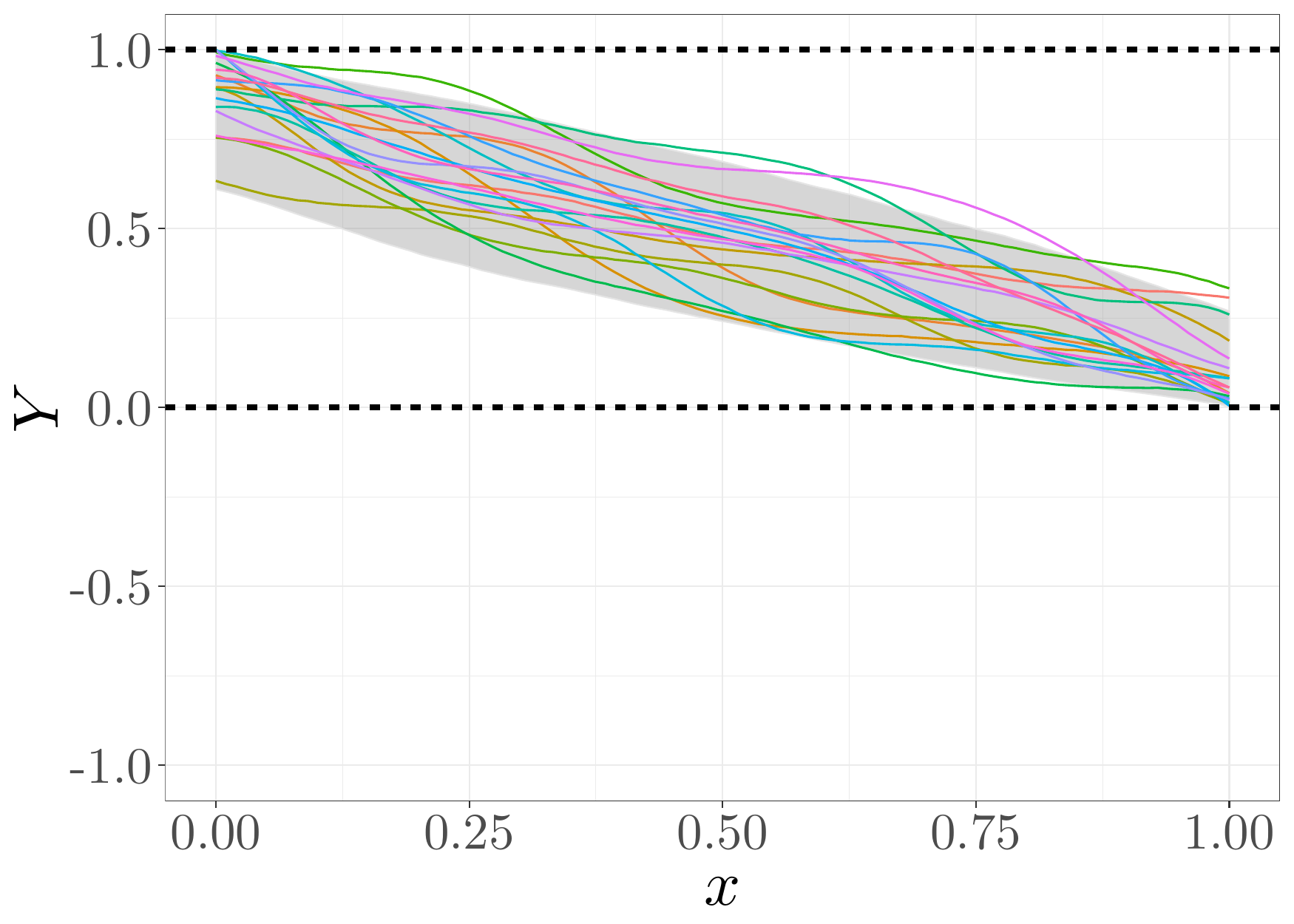}
	
	\includegraphics[width=0.325\textwidth]{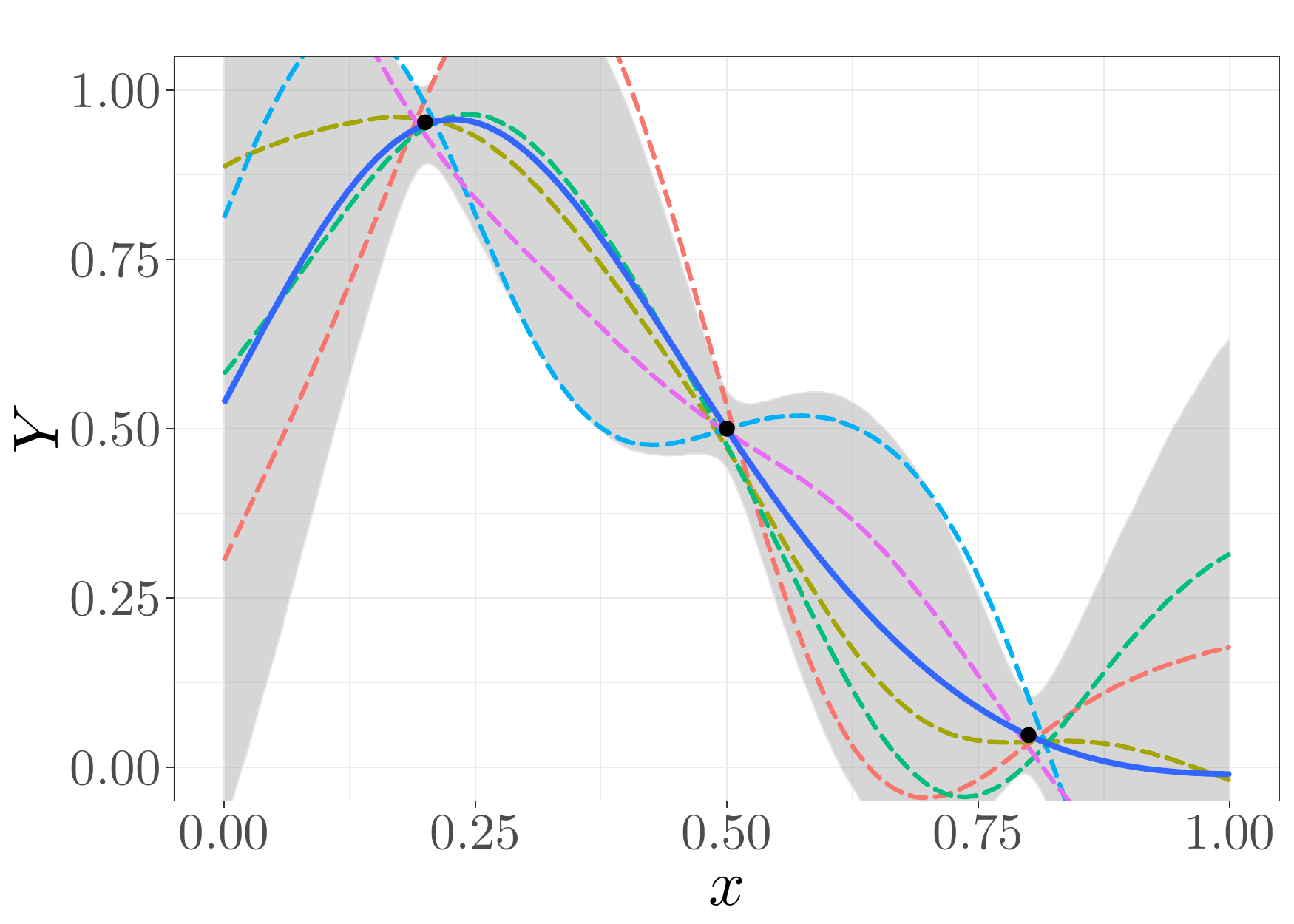}
	\includegraphics[width=0.325\textwidth]{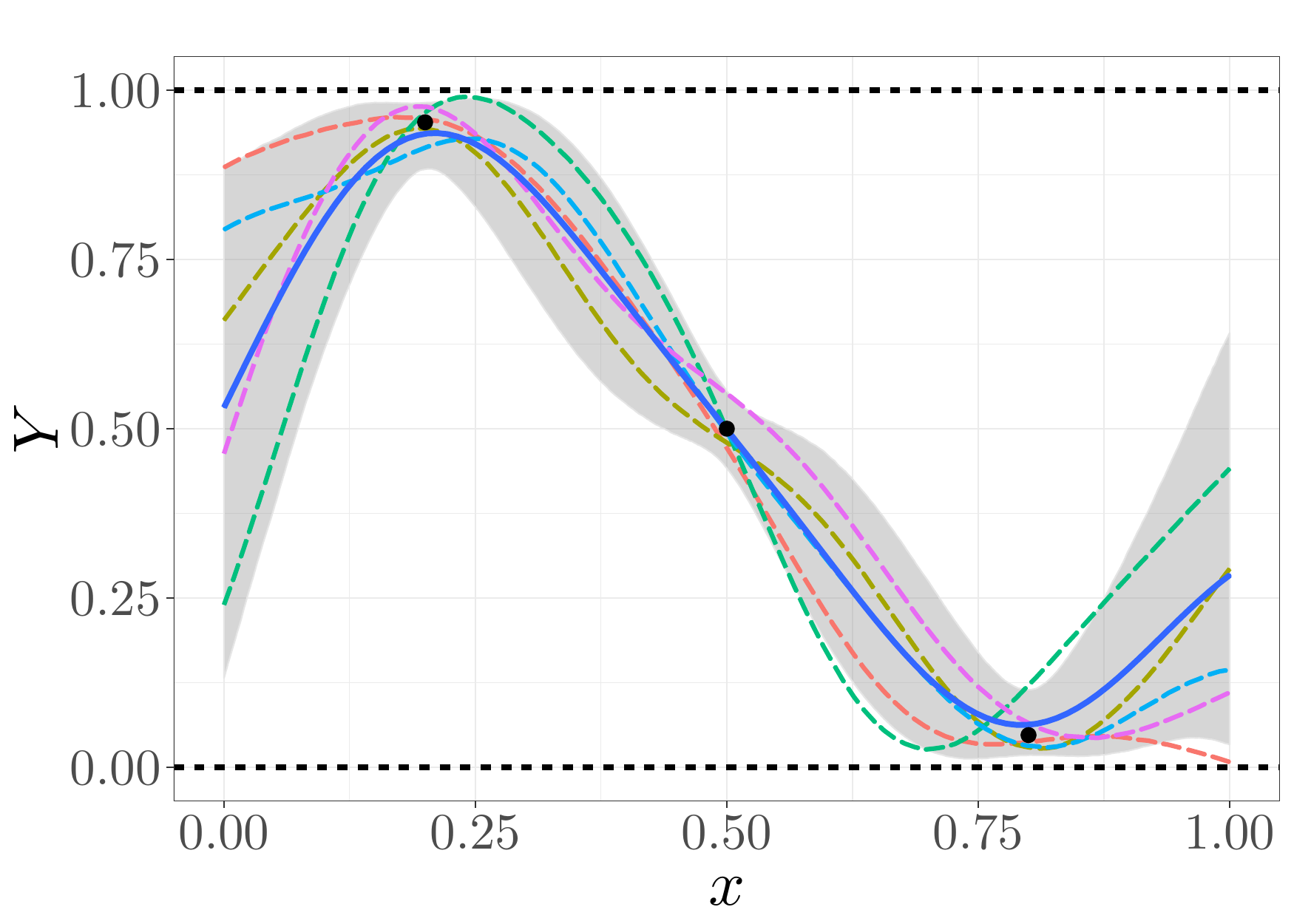}
	\includegraphics[width=0.325\textwidth]{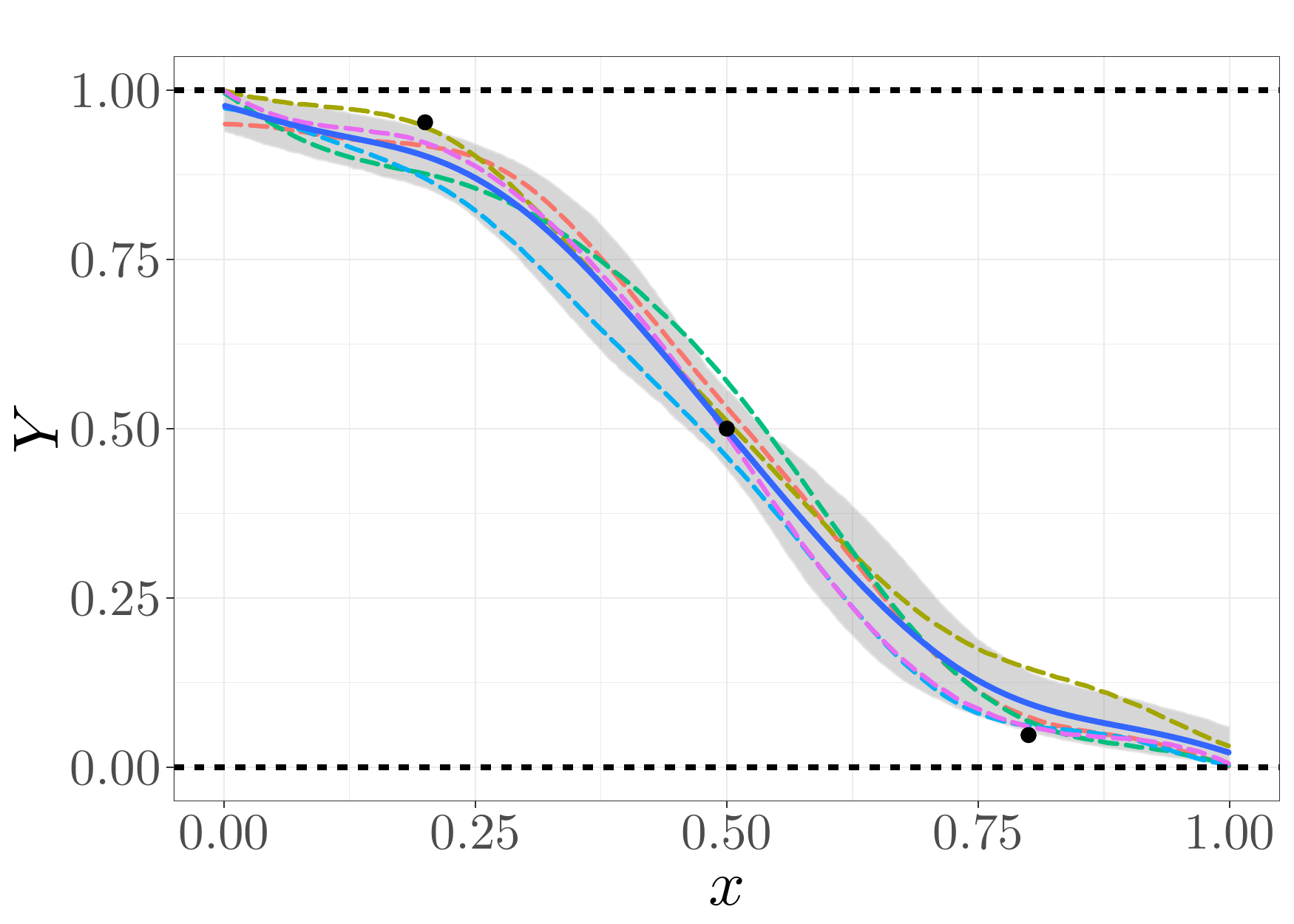}
	\caption{GP emulators under no constraints (left), boundedness constraints $Y \in [0,1]$ (centre), boundedness $Y \in [0,1]$ and non-increasing constraints (right). Samples from the different types of (constrained) Gaussian priors and the resulting GP emulators are shown in the first and second row, respectively. Each panel shows: the conditional emulations (dashed lines), and the 95\% prediction interval (grey region). For boundedness constraints, bounds at $l = 0$ and $u = 1$ correspond to horizontal dashed lines. For GP emulators, the conditional mean (blue solid line) and interpolation points (dots) are also shown.}
	\label{fig:lineqGPexamples}
\end{figure*}

In this paper, we aim at imposing linear inequality constraints on Gaussian process (GP) emulators when observations are considered noisy. As an example, Figure~\ref{fig:lineqGPexamples} shows three types of GP emulators $Y$ with training points at $x_1 = 0.2$, $x_2 = 0.5$, $x_3 = 0.8$, and different inequality conditions. We used a squared exponential (SE) covariance function,
\begin{equation*}
k_{\Btheta}(x,x') = \sigma^2 \exp\left\{-\frac{(x-x')^2}{2\ell^2} \right\},
\end{equation*}
with $\Btheta = (\sigma^2, \ell)$. We fixed the variance parameter $\sigma^2 = 0.5^2$ and length-scale parameter $\ell = 0.2$. We set a noise variance to be equal to 0.5\% of the variance parameter $\sigma^2$. One can observe that different types of (constrained) Gaussian priors (top) yield different GP emulators (bottom) for the same training data. One can also note that the interpolation constraints are relaxed due to the noise effect, and that the inequality constraints are still satisfied everywhere.

Next, we formally introduce the corresponding model to obtain constrained GP emulators with linear inequality constraints and noisy observations as in Figure~\ref{fig:lineqGPexamples}.

\subsection{Finite-Dimensional Approximation of Gaussian Process Emulators with Noisy Observations}
\label{subsec:finitContrGPs}
Let $Y$ be a centred GP on $\realset{}$ with arbitrary covariance function $k$. Consider $x \in \mathcal{D}$, with space $\mathcal{D} = [0, 1]$. For simplicity, consider a spline decomposition with an equispaced set of knots $t_1, \cdots, t_m \in \mathcal{D}$ such that $t_j = (j-1) \Delta_m$ for $j = 1, \cdots, m$, with $\Delta_m = 1/(m-1)$. This assumption can be relaxed for non-equispaced designs of knots as in \citep{Larson2013FEM}, leading to similar developments as obtained in this paper but with slight differences when imposing the inequality constraints (e.g. convexity condition). 
In contrast to \citep{LopezLopera2017FiniteGPlinear}, here we consider noisy observations $y_i \in \realset{}$ for $i = 1, \cdots, n$. Define $Y_m$ as a stochastic emulator consisting of the piecewise-linear interpolation of $Y$ at knots $(t_1, \cdots, t_m)$,
\begin{equation}
Y_m (x) = \sum_{j=1}^{m} \phi_j (x) Y(t_j), \quad \mbox{s.t.} \quad Y_m (x_i) + \varepsilon_i = y_i \ \mbox{(interpolation constraints)},
\label{eq:finApproxGPEmulator}
\end{equation}
where $x_i \in \mathcal{D}$, $\varepsilon_i \sim \normrnd{0}{\tau^2}$ for $i = 1, \cdots, n$, with noise variance $\tau^2$, and $\phi_1, \cdots, \phi_m$ are hat basis functions given by
\begin{equation}
\phi_j (x) :=
\begin{cases}
1 - \left|\frac{x - t_j}{\Delta_m}\right| & \mbox{if } \left|\frac{x - t_j}{\Delta_m}\right| \leq 1,\\
0 & \mbox{otherwise}.
\end{cases}
\label{eq:hatbasisfun}
\end{equation}
As in many classical GP implementations \citep{Rasmussen2005GP}, we assume that $\varepsilon_1, \cdots, \varepsilon_n$ are independent, and independent of $Y$. However, since the framework proposed here does not have any restriction on the type of the covariance function, the extension to other noise distributions and/or noise with autocorrelation can be done as in standard GP implementations \citep{Rasmussen2005GP,Murphy2012ML}.

One must note that the benefit of considering noisy observations in~\eqref{eq:finApproxGPEmulator} is that, due to the ``relaxation'' of the interpolation conditions, the number of knots $m$ does not have to be larger than the number of interpolation points $n$ (assumption required in \citep{LopezLopera2017FiniteGPlinear} for the interpolation of noise-free observations). Then, for $m \ll n$, the finite representation in~\eqref{eq:finApproxGPEmulator} would lead to less expensive procedures since the cost of the MC and MCMC procedures below grow with the value of $m$ rather than $n$ (see Sections~\ref{subsec:contrGPsPosterior} and~\ref{subsec:MAP1D}).

\subsection{Imposing Linear Inequality Constraints}
\label{subsec:contrGPsPosterior}
Now, assume that $Y_m$ also satisfies inequality constraints everywhere in the input space (e.g. boundedness, monotonicity, convexity), i.e.
\begin{equation}
Y_m \in \mathcal{E} \quad \mbox{(inequality constraints)},
\label{eq:constrFinApproxGPEmulator}		
\end{equation}
where $\mathcal{E}$ is a convex set of functions defined by some inequality conditions. Then, the benefit of using~\eqref{eq:finApproxGPEmulator} is that, for many constraint sets $\mathcal{E}$, satisfying $Y_m \in \mathcal{E}$ is equivalent to satisfying only a finite number of inequality constraints at the knots $(Y(t_1), \cdots, Y(t_m))$ \citep{Maatouk2017GPineqconst}, i.e.
\begin{equation}
Y_m \in \mathcal{E} \; \; \Leftrightarrow \; \; \Bxi \in \mathcal{C},
\label{eq:convexEqui}
\end{equation}
with $\xi_j := Y(t_j)$ for $j = 1, \cdots, m$, and $\mathcal{C}$ a convex set on $\realset{m}$. As an example, when we evaluate a GP with bounded trajectories $l \leq Y_m (x) \leq u$, the convex set $\mathcal{C}$ can be defined by $\mathcal{C}_{[l,u]} := \{ \textbf{c} \in \realset{m}; \ \forall \ j = 1, \cdots, m \ : l \leq \ c_j \leq u \}$.
In this paper, we consider the case where $\mathcal{C}$ is composed by a set of $q$ linear inequalities of the form
\begin{equation}
\mathcal{C} = \bigg\{\textbf{c} \in \realset{m}; \ \forall \; k = 1, \dots, q \; : \; l_k \leq \sum_{j = 1}^{m} \lambda_{k,j} c_j \leq u_k \bigg\},
\label{eq:convexsetKnotsLinear}	
\end{equation}
where the $\lambda_{k,j}$'s encode the linear operations, the $l_k$'s and $u_k$'s represent the lower and upper bounds, respectively. One can note that the convex set $\mathcal{C}_{[l,u]}$ is a particular case of $\mathcal{C}$ where $\lambda_{k,j} = 1$ if $k = j$ and zero otherwise, and with bounds $l_k = l$, $u_k = u$, for $k = 1, \cdots, m$. 

We now aim at computing the distribution of $Y_m$ conditionally on the constraints in~\eqref{eq:finApproxGPEmulator} and~\eqref{eq:constrFinApproxGPEmulator}. One can observe that the vector $\Bxi$ is a centred Gaussian vector with covariance matrix $\BGamma= (k(t_i,t_j))_{1 \leq i,j \leq m}$. Denote $\BLambda = (\lambda_{k,j})_{1 \leq k \leq q, 1 \leq j \leq m}$, $\Bl = (\ell_{k})_{1 \leq k \leq q}$, $\Bu = (u_{k})_{1 \leq k \leq q}$, $\BPhi$ the $n \times m$ matrix defined by $\BPhi_{i,j} = \phi_j (x_i)$, and $\By = [y_1, \cdots,  y_n]^\top$ the vector of noisy observations at points $x_1 , \cdots, x_n$. Then, the distribution of $\Bxi$ conditioned on $\BPhi \Bxi + \Bvarepsilon= \By$, with $\Bvarepsilon \sim \normrnd{\Bzero}{\tau^2 \Beye}$, is given by \citep{Rasmussen2005GP}
\begin{equation}
\Bxi |\{\BPhi \Bxi + \Bvarepsilon = \By\} \sim \mathcal{N} (\Bmu, \BSigma),
\label{eq:posterior}
\end{equation}
where
\begin{equation}
\Bmu = \BGamma \BPhi^\top [\BPhi \BGamma \BPhi^\top + \tau^2 \Beye]^{-1} \By, \quad \mbox{and} \quad 
\BSigma = \BGamma -  \BGamma \BPhi^\top [\BPhi \BGamma \BPhi^\top + \tau^2 \Beye]^{-1} \BPhi \BGamma.
\label{eq:posteriorPar}
\end{equation}
One can note that, in the limit as the noise variance $\tau^2 \to \infty$, then $\Bmu \to \Bzero$ and $\BSigma \to \BGamma$, and therefore the distribution in~\eqref{eq:posterior} ignores the observations $\By$. In that case, MC and MCMC samplers are performed in the sample space of the prior of $\Bxi$, which is less restrictive than the one of $\Bxi |\{\BPhi \Bxi + \Bvarepsilon = \By\}$. Since the inequality constraints are on $\BLambda \Bxi$, one can first show that the posterior distribution of $\BLambda \Bxi$ conditioned on $\BPhi \Bxi + \Bvarepsilon = \By$ and $\Bl \leq \BLambda \Bxi \leq \Bu$ is truncated Gaussian-distributed \citep[see, e.g.,][for further discussion when noise-free observations are considered]{LopezLopera2017FiniteGPlinear}, i.e.	
\begin{equation}
\BLambda \Bxi|\{\BPhi \Bxi + \Bvarepsilon = \By, \Bl \leq \BLambda \Bxi \leq \Bu\} \sim \tnormrnd{\BLambda \Bmu}{\; \BLambda \BSigma \BLambda^\top}{\; \Bl}{\; \Bu}.
\label{eq:trposterior}
\end{equation}
Notice that the inequality constraints are encoded in the posterior mean $\BLambda \Bmu$, the posterior covariance $\BLambda \BSigma \BLambda^\top$, and the bounds $(\Bl,\Bu)$. Moreover, one must also highlight that by considering noisy observations, due to the ``relaxation'' of the interpolation conditions, inequality constraints can be imposed also when the observations $(y_1, \cdots, y_n)$ do not fulfil the inequalities. 

Finally, the truncated Gaussian distribution in~\eqref{eq:trposterior} does not have a closed-form expression but it can be approximated via MC or MCMC. Hence, samples of $\Bxi$ can be recovered from samples of $\BLambda \Bxi$, by solving a linear system \citep{LopezLopera2017FiniteGPlinear}. As discussed in \citep{LopezLopera2017FiniteGPlinear}, the number of inequalities $q$ is usually larger than the number of knots $m$ for many convex sets $\mathcal{C}$. If we further assume that $q \geq  m$, and that $\operatorname{rank}(\BLambda) = m$, then the solution of the linear system $\BLambda \Bxi$ exists and is unique \citep[see][for a further discussion]{LopezLopera2017FiniteGPlinear}. Therefore, samples of $Y_m$ can be obtained from samples of $\Bxi$, with the formula $Y_m (x) = \sum_{j=1}^{m} \phi_j (x) \xi_j$ for $x \in \mathcal{D}$. The implementation of the GP emulator $Y_m$ is summarised in Algorithm~\ref{alg:tmKriging}.
\begin{algorithm}[t!]
	\small	
	\caption{GP emulator with linear inequality constraints.}\label{alg:tmKriging}
	\begin{algorithmic}[1]
		\State \textbf{REQUIRE:} $\By \in \realset{n}$, $\BGamma \in \realset{m\times m}$, $\tau^2 \in \realset{+}$, $\BPhi \in \realset{n\times m}$, $\BLambda \in \realset{q \times m}$, $\Bl \in \realset{q}$, $\Bu \in \realset{q}$
		\State \textbf{ENSURE:} Emulated samples from $\Bxi |\{ \BPhi \Bxi + \Bvarepsilon= \By, \Bl \leq \BLambda \Bxi \leq \Bu \}$
		\State Compute the conditional mean and covariance of $\Bxi |\{ \BPhi \Bxi + \Bvarepsilon = \By \}$,
		\State \hspace{10pt} $\Bmu = \BGamma \BPhi^\top (\BPhi \BGamma \BPhi^\top + \tau^2 \Beye)^{-1} \By$,
		\State \hspace{10pt} $\BSigma = \BGamma - \BGamma \BPhi^\top (\BPhi \BGamma \BPhi^\top + \tau^2 \Beye)^{-1} \BPhi \BGamma$.
		\State Sample $\Bz$ from the truncated Gaussian distribution via MC/MCMC, 
		\State \hspace{10pt} $\Bz = \BLambda \Bxi |\{ \BPhi \Bxi + \Bvarepsilon = \By, \textbf{\textit{l}} \leq \BLambda \Bxi \leq \textbf{\textit{u}} \} \sim \tnormrnd{\BLambda \Bmu}{\; \BLambda \BSigma\BLambda^\top}{\; \Bl}{\; \Bu}.$
		\State Compute $\Bxi$ by solving the linear system $\BLambda \Bxi = \Bz$.
	\end{algorithmic}
\end{algorithm}

\subsection{Maximum a Posteriori Estimate via Quadratic Programming}
\label{subsec:MAP1D}
In practice, the posterior mode (maximum a posteriori estimate, MAP) of~\eqref{eq:trposterior} can be used as a point estimate of unobserved quantities \citep{Rasmussen2005GP}, and as a starting state of MCMC samplers \citep{Murphy2012ML}. Let $\Bmu^\ast$ be the posterior mode that maximises the probability density function (pdf) of $\Bxi$ conditioned on $\BPhi \Bxi + \Bvarepsilon = \By$ and $\Bl \leq \BLambda \Bxi \leq \Bu$.  Then, maximising the pdf in~\eqref{eq:trposterior} is equivalent to maximise the quadratic problem
\begin{equation}
\Bmu^\ast = \underset{\Bxi \text{ s.t. } \Bl \leq \BLambda \Bxi \leq \Bu}{\arg\max} \{- [\Bxi - \Bmu]^{\top} \BSigma^{-1} [\Bxi - \Bmu]\},
\label{eq:MAP1}
\end{equation}
with conditional mean $\Bmu$ and conditional covariance $\BSigma$ as in~\eqref{eq:posteriorPar}. By maximising~\eqref{eq:MAP1}, we are looking for the most likely vector $\Bxi$ satisfying both the interpolation and inequality constraints \citep{Bishop2007ML}. One must highlight that the posterior mode of~\eqref{eq:trposterior} converges uniformly to the spline solution when the number of knots $m \to \infty$ \citep{Maatouk2017GPineqconst,Bay2016KimeldorfWahba}. Finally, the optimisation problem in~\eqref{eq:MAP1} is equivalent to
\begin{equation}
\Bmu^\ast
=  \underset{\Bxi \text{ s.t. } \Bl \leq \BLambda \Bxi \leq \Bu}{\arg\min} \{\Bxi^\top \BSigma^{-1} \Bxi - 2\Bmu^{\top} \BSigma^{-1} \Bxi \},
\label{eq:MAP2}	
\end{equation}
which can be solved via quadratic programming \citep{Goldfarb1982QP}.

\subsection{Extension to Higher Dimensions}
\label{subsec:contrGPsHD}
The GP emulator of Section~\ref{sec:contrGPs} can be extended to $d$ dimensional input spaces by tensorisation \citep[see, e.g.,][for a further discussion on imposing inequality constraints for $d \geq 2$]{Maatouk2017GPineqconst,LopezLopera2017FiniteGPlinear}. Consider $\Bx = (x_1, \cdots, x_d) \in \mathcal{D}$ with input space $\mathcal{D} = [0, 1]^d$, and a set of knots per dimension $(t_1^1, \cdots, t_{m_1}^1), \cdots, (t_1^d, \cdots, t_{m_d}^d)$. Then, the GP emulator $Y_{m_1,\cdots,m_d}$ is given by
\begin{equation}
Y_{m_1,\cdots,m_d} (\Bx) = \sum_{j_1= 1, \cdots, m_1} \cdots \sum_{j_d= 1, \cdots, m_d} [\phi_{j_1}^1 (x_1) \times \cdots \times \phi_{j_d}^d (x_d)] \xi_{j_1, \cdots, j_d},
\label{eq:finApproxHD}
\end{equation}	
with $\Bx_i \in \mathcal{D}$ for $i = 1, \cdots, n$, $\xi_{j_1, \cdots, j_d} := Y(t_{j_1}, \cdots, t_{j_d})$, and $\phi_{j_\kappa}^\kappa$ are hat basis functions as defined in~\eqref{eq:hatbasisfun}. We aim at computing \eqref{eq:finApproxHD} subject to some interpolation constraints $Y_{m_1,\cdots,m_d} \left(\Bx_{i}\right) + \varepsilon_i = y_i$, with $y_i \in \realset{}$ and $\varepsilon_i \sim \normrnd{0}{\tau^2}$ for $i = 1, \cdots, n$; and inequality constraints $\Bxi = [\xi_{1, \cdots,1}, \cdots, \xi_{m_1, \cdots, m_d}]^\top \in \mathcal{C}$ with $\mathcal{C}$ a convex set of $\realset{m_1 \times \cdots \times m_d}$. We assume that $\varepsilon_1, \cdots, \varepsilon_n$ are independent, independent of $Y$.	Then, following a similar procedure as in Section~\ref{sec:contrGPs}, Algorithm~\ref{alg:tmKriging} can be used with $\Bxi$ a centred Gaussian vector with an arbitrary covariance matrix $\BGamma$. 

Notice that having less knots than observations can have a great impact since the MC and MCMC samplers will then be performed in low dimensional spaces when $m = m_1 \times \cdots \times m_d \ll n$. For the case $m \ll n$, the inversion of the matrix $(\BPhi \BGamma \BPhi^\top + \tau^2 \Beye) \in \realset{n \times n}$ can be computed more efficiently through the matrix inversion lemma \citep{Press1992Num}, reducing the computational complexity to the inversion of an $m \times m$ full-rank matrix. Therefore, the computation of the conditional distribution in~\eqref{eq:posteriorPar} and the estimation of the covariance parameter can be achieved faster. Moreover, due to the relaxation of the interpolation conditions through a noise effect, MC and MCMC samplers are performed in less restrictive sample spaces, and this leads to faster emulators.

\section{Numerical Experiments}
\label{sec:numExperiments}

The codes were implemented in the R programming language, based on the open source package \texttt{lineqGPR} \citep{LopezLopera2018LineqGPR}. This package is based on previous R software developments produced by the Dice (Deep Inside Computer Experiments) and ReDice Consortiums (e.g. DiceKriging, \citet{Roustant2012DiceKriging}; DiceDesign, \citet{Dupuy2015DiceDesign}; kergp \citet{Deville2015kergp}), but incorporating some structures of classic libraries for GP regression modelling from other platforms (e.g. the GPmat toolbox from MATLAB, and the GPy library from Python).

\texttt{lineqGPR} also contains implementations of different samplers for the approximation of truncated (multivariate) Gaussian distribution. Samplers are based on recent contributions on efficient MC and MCMC inference methods. Table~\ref{tab:comparissonSamplers} summarise some properties of the different MC and MCMC samplers used in this paper \citep[see, e.g.,][for a further discussion]{Maatouk2016RSM,Botev2017MinimaxTilting,Benjamini2017fastGibbs,Pakman2014Hamiltonian}.
\begin{table}[t!]
	\centering
	\caption{Comparison between the different MC and MCMC samplers provided in \texttt{lineqGPR}: rejection sampling from the mode (RSM) \citep{Maatouk2016RSM}, exponential tilting (ExpT) \citep{Botev2017MinimaxTilting}, Gibbs sampling \citep{Benjamini2017fastGibbs}, and exact Hamiltonian monte carlo (HMC)\citep{Pakman2014Hamiltonian}.}
	\label{tab:comparissonSamplers}
	\begin{tabular}{ccccc}
		\toprule						
		Item & RSM & ExpT & Gibbs & HMC \\
		\midrule
		Exact method & \goodcheck & \goodcheck & \badcheck & \badcheck \\
		Non parametric & \goodcheck & \goodcheck &  \goodcheck & \goodcheck\\ 
		Acceptance rate & low & high & 100\% & 100\% \\
		Speed & slow & fast & slow-fast & fast \\
		Uncorrelated samples & \goodcheck & \goodcheck & \badcheck & \badcheck\\ 
		Previous R Implementations & \texttt{constrKriging} & \texttt{TruncatedNormal} & \texttt{tmvtnorm} & \texttt{tmg}\\
		\bottomrule
	\end{tabular}
\end{table}

Codes were executed on a single core of an Intel\textsuperscript{\textregistered} Core\textsuperscript{TM} i7-6700HQ CPU.

\subsection{1D Toy Example under Boundedness Constraint}
\label{subsec:toy1DBoundedness}
Here, we use the GP framework introduced in Section~\ref{sec:contrGPs} for emulating bounded trajectories $Y_m \in [-\alpha, \alpha]$ with constant $\alpha \in \realset{}$. We aim at analysing the resulting constrained GP emulator when noise-free or noisy observations are considered. The dataset is $(x_i,y_i)_{1 \leq i \leq 5}$: $(0, 0)$, $(0.2, -0.5)$, $(0.5, -0.3)$, $(0.75, 0.5)$, and $(1, 0.4)$. We use a Mat\'ern 5/2 covariance function,
\begin{equation*}
k_{\Btheta}(x,x') = \sigma^2 \left(1 + \frac{\sqrt{5} |x-x'|}{\ell} + \frac{5}{3} \frac{(x-x')^2}{\ell^2}\right) \exp\left\{-\frac{\sqrt{5}|x-x'|}{\ell} \right\},
\end{equation*}
with $\Btheta = (\sigma^2, \ell)$. We fix the variance parameter $\sigma^2 = 10$ leading to highly variable trajectories. The lengthscale parameter $\ell$ and the noise variance $\tau^2$ are estimated via maximum likelihood (ML).

\begin{figure}[t!]
	\centering
	\includegraphics[width=0.325\textwidth]{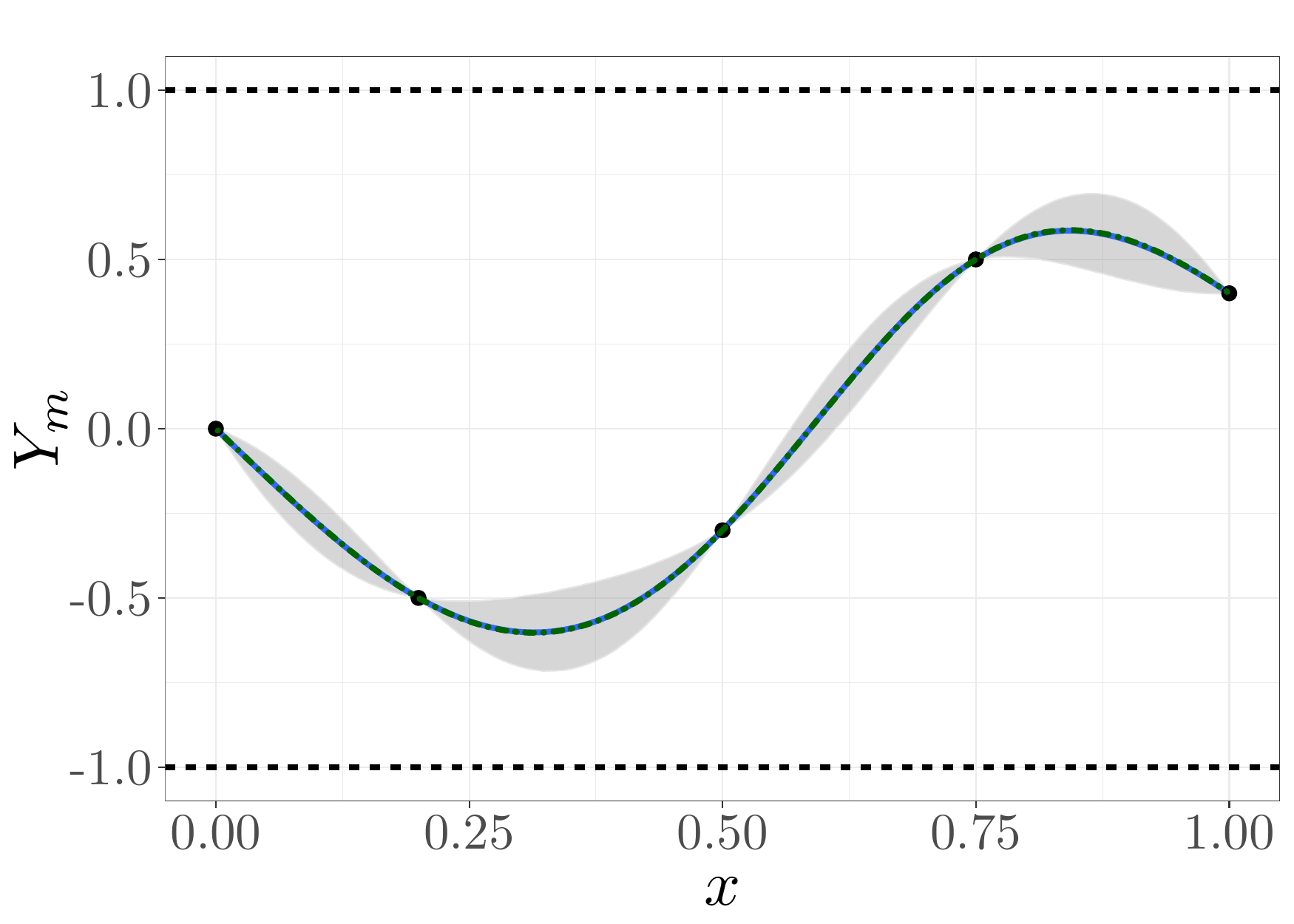}
	\includegraphics[width=0.325\textwidth]{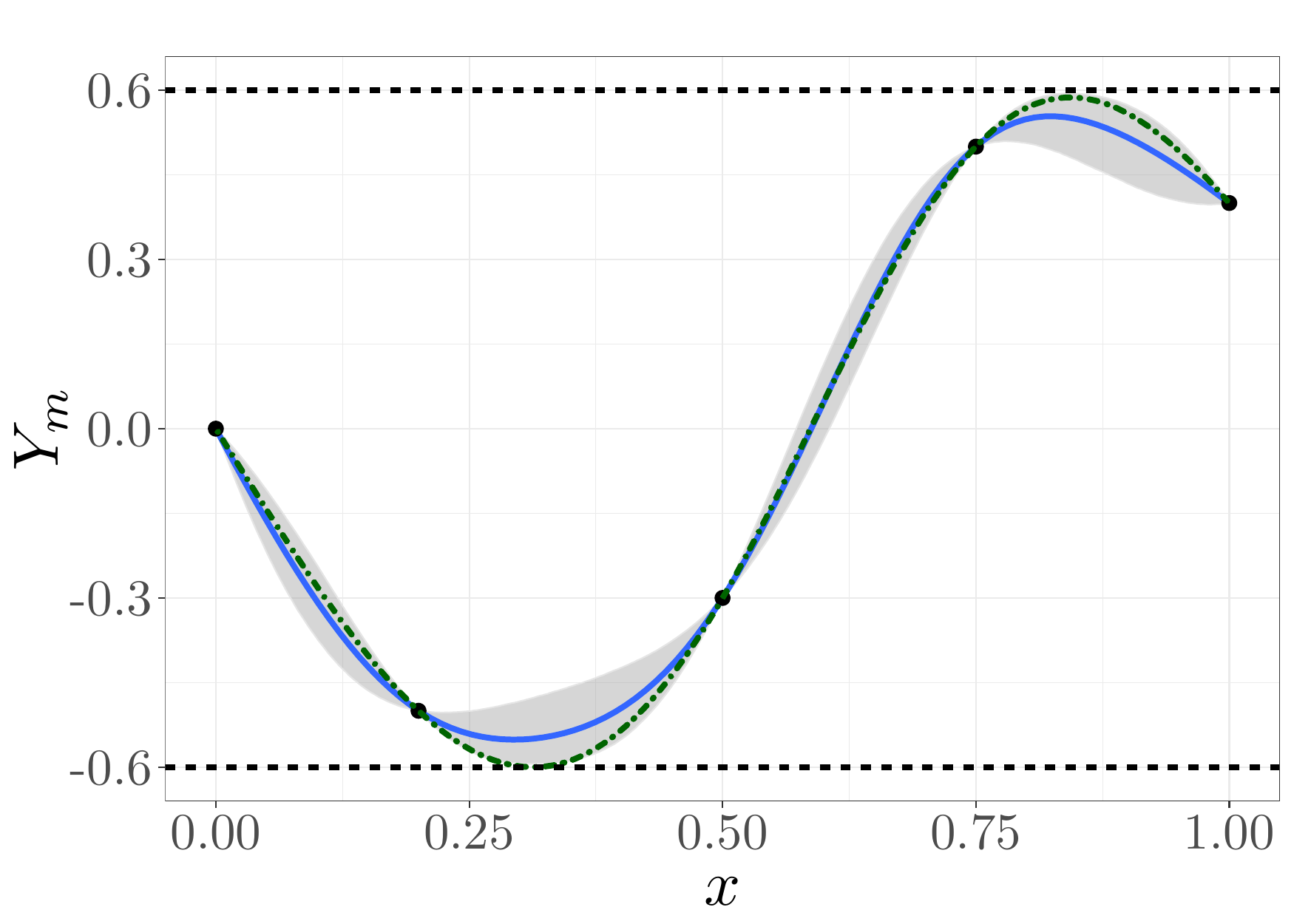}
	\includegraphics[width=0.325\textwidth]{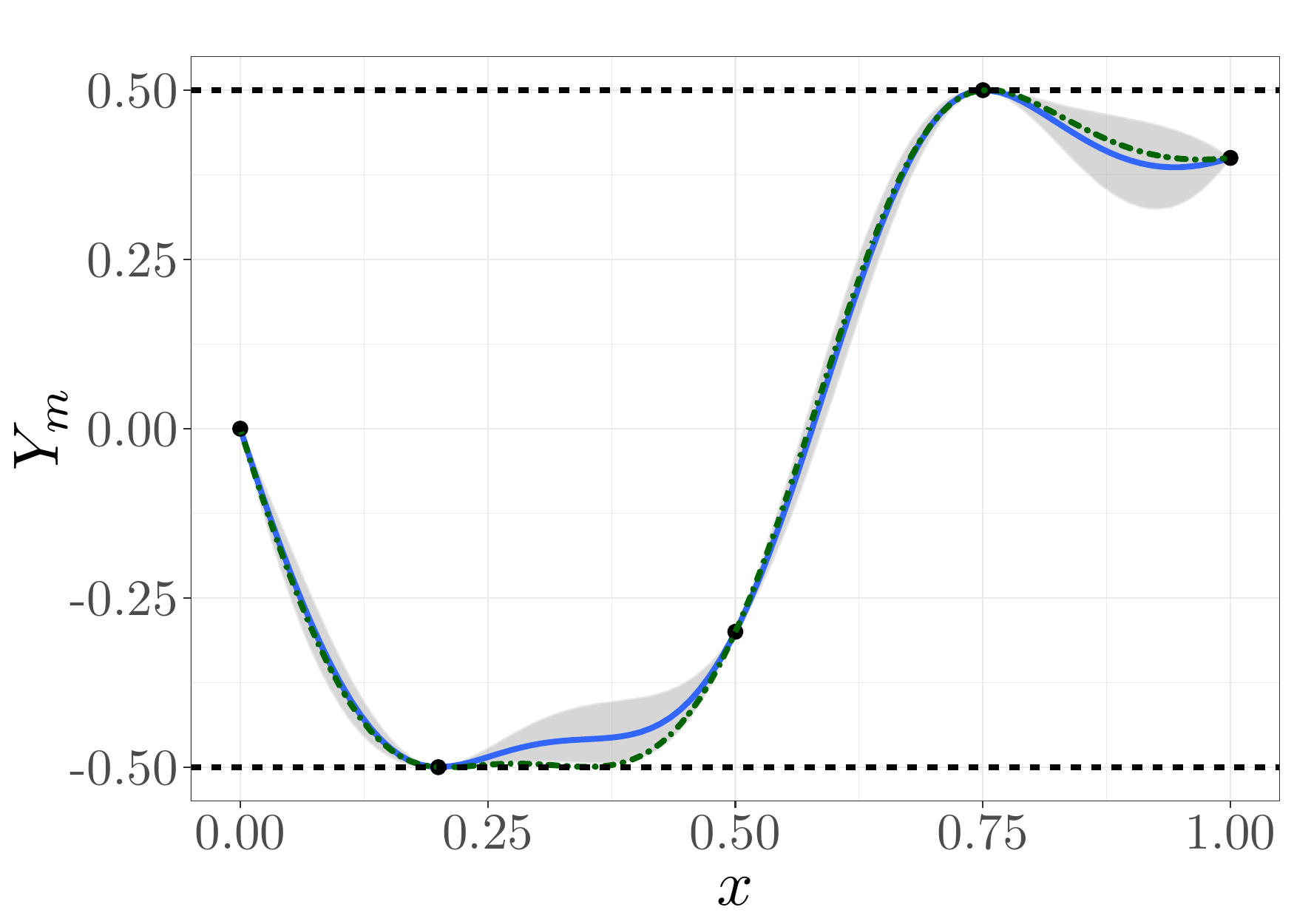}
	
	\includegraphics[width=0.325\textwidth]{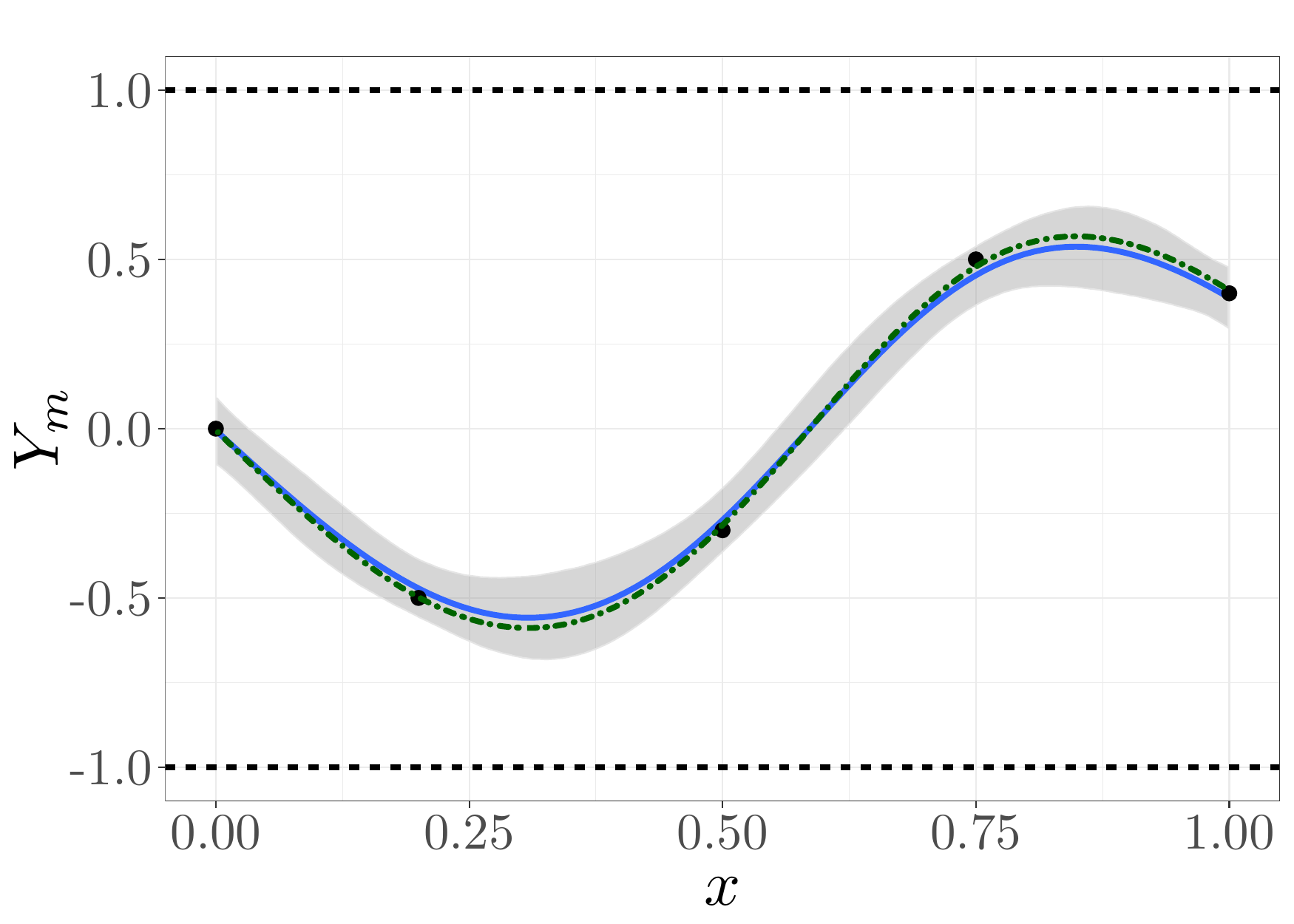}
	\includegraphics[width=0.325\textwidth]{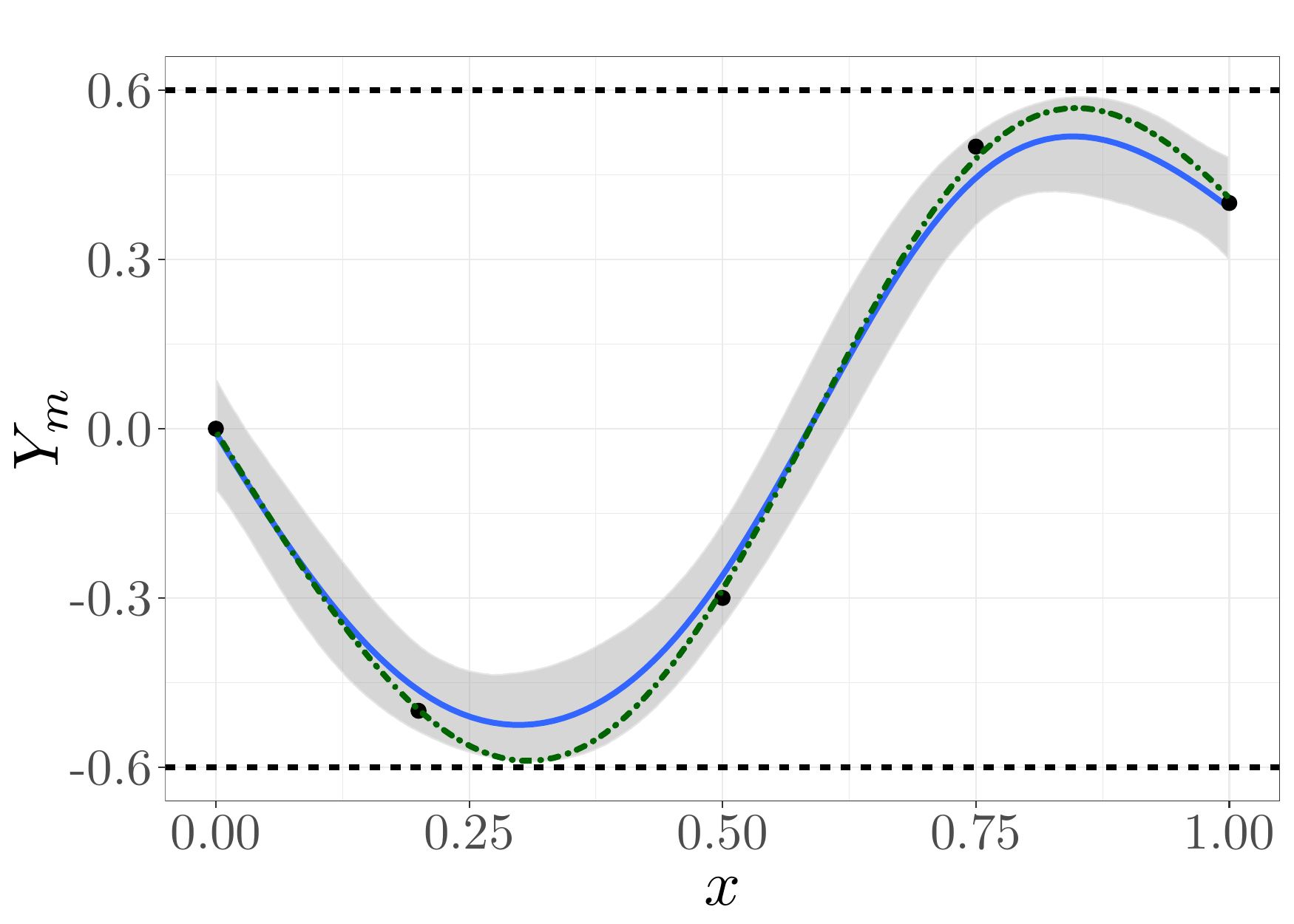}
	\includegraphics[width=0.325\textwidth]{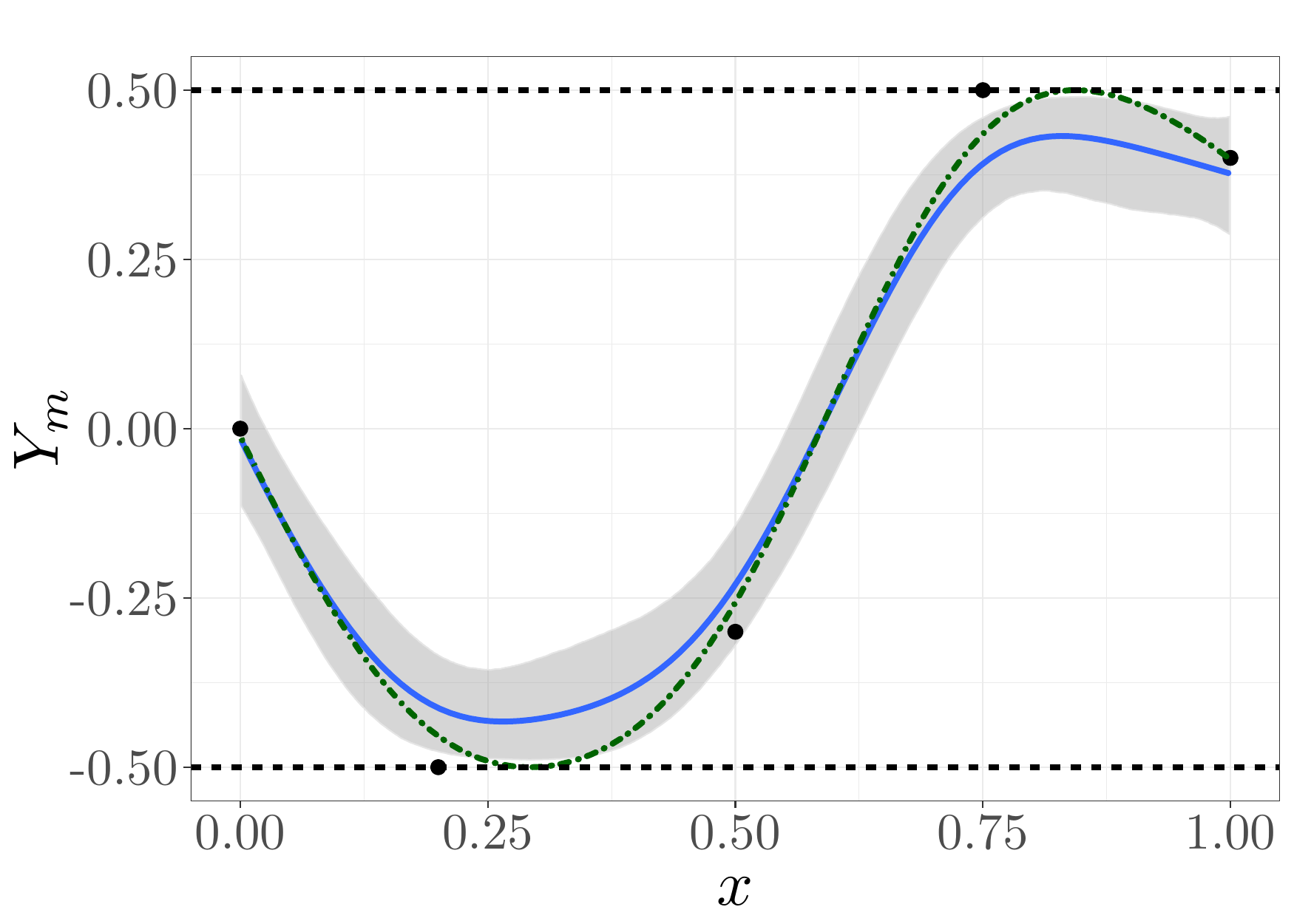}
	
	\caption{GP emulators under boundedness constraints $Y_m \in [-\alpha, \alpha]$. Results are shown considering (top) noise-free and (bottom) noisy observations: (left) $\alpha = 1$, (centre) $\alpha = 0.6$, and (right) $\alpha = 0.5$.  Each panel shows: the observations (dots), the conditional mean (solid line), the conditional mode (dot-dash line), the 95\% prediction interval (grey region), and the bounds (dashed lines).}
	\label{fig:Comparisson}
\end{figure}

The effect of different bounds $[-\alpha, \alpha]$ on the constrained GP emulators can be seen in Figure~\ref{fig:Comparisson}. There, we set $m = 100$ for having emulations with high-quality of resolution, and we generated $10^4$ constrained emulations via RSM \citep{Maatouk2016RSM}. One can observe that, since interpolation conditions were relaxed due to the influence of the noise variance $\tau^2$, the prediction intervals are wider when bounds become closer to the observations. For the case $\alpha = 0.5$, the noise-free GP emulator yielded costly procedures due to a small acceptance rate equal to $0.1\%$. In contrast, when noisy observations were assumed, emulations were more likely to be accepted leading to an acceptance rate equal to $16.92\%$.

Now, we assess all the MC and MCMC methods from Table~\ref{tab:comparissonSamplers} for the approximation of the truncated Gaussian posterior distribution in~\eqref{eq:trposterior}. We considered the examples from Figure~\ref{fig:Comparisson}. For the MCMC samplers, we used the posterior mode solution from~\eqref{eq:MAP2} as the starting state of the Markov chains. This initialises the chains in a high probability region. Therefore, only few emulations have been ``burned'' in order to have samples that appeared to be independent of the starting state. Here, we only burned the first 100 emulations. We evaluated the performance of both MC and MCMC samplers in terms of the effective sample size (ESS):
\begin{equation}
\operatorname{ESS} = \frac{n_s}{1+2\sum_{k=1}^{n_s} \rho_k},
\label{eq:ESS}
\end{equation}	
where $n_s$ is the size of the sample path, and $\rho_k$ is the sample autocorrelation with lag $k$. The ESS indicator gives an intuition on how many emulations of the sample path can be considered independent \citep{Gong2016ESS}. In order to obtain non-negative sample autocorrelations $\rho_k$, we used the convex sequence estimator proposed in \citep{Geyer1992MCMC}. We then computed the ESS of each coordinate of $\Bxi \in \realset{m}$, i.e. $\operatorname{ESS}_j = \operatorname{ESS}(\xi_j^1, \cdots, \xi_j^{n_s})$ for $j = 1, \cdots, m$, and we evaluated the quantiles $(q_{10\%},q_{50\%},q_{90\%})$ over the $m$ resulting $\operatorname{ESS}$ values. The sample size $n_s = 10^4$ has been chosen to be larger than the minimum ESS required to obtain a proper estimation of the vector $\Bxi \in \realset{m}$ \citep{Gong2016ESS}. Finally, we tested the efficiency of each sampler by computing the time normalised ESS (TN-ESS)  \citep{Lan2016Sampling} at $q_{10\%}$ (worst case): $\operatorname{TN-ESS} = q_{10\%}(\operatorname{ESS})/\operatorname{(CPU \ Time)}$.

Table~\ref{tab:comparisson_m100} displays the performance indicators obtained for each samplers from Table~\ref{tab:comparissonSamplers}. Firstly, one can observe that RSM yielded the most expensive procedures due to its high rejection rate when sampling the constrained trajectories from the posterior mode. In particular, for $\alpha = 0.5$, and assuming noise-free observations, the prohibitively small acceptance rate of RSM led to costly procedures (about 7 hours) making it impractical. Secondly, although the Gibbs sampler needs to discard intermediate samples (thinning effect), it provided accurate ESS values within a moderate running time (with effective sampling rates of $400 \ s^{-1}$). Thirdly, due to the high acceptance rates obtained by ExpT, and good exploratory behaviour of the exact HMC, both samplers provided much more efficient TN-ESS values compared to their competitors, generating thousands of effective emulations each second. Finally, as we expected, the performance of some samplers were improved when adding a noise. For RSM, due to the relaxation of the interpolation conditions, we noted that emulations were more likely to be accepted leading quicker routines: more than 150 times faster with noise (see Table~\ref{tab:comparisson_m100}, $\alpha = 0.5$).	
\begin{table}[t!]
	\centering	
	\footnotesize
	\caption{Efficiency of MC and MCMC from Table~\ref{tab:comparissonSamplers} for emulating bounded samples $Y_m \in [-\alpha,\alpha]$ of Figure~\ref{fig:Comparisson}. Best results are shown in bold. For the Gibbs sampler, we set the thinning parameter to 200 emulations aiming to obtain competitive ESS values with respect to other samplers. $^\dagger$Results could not be obtained due to numerical instabilities.}
	\label{tab:comparisson_m100}
	\begin{tabular}{c|c|ccc|ccc}
		\toprule
		\multirow{3}{*}{Bounds} & \multirow{3}{*}{Method} & \multicolumn{3}{c|}{Without noise variance} & \multicolumn{3}{c}{With noise variance} \\
		& & CPU Time & ESS $[\times 10^4]$ & TN-ESS & CPU Time & ESS $[\times 10^4]$ & TN-ESS \\
		& & $[s]$ & $(q_{10\%},q_{50\%},q_{90\%})$ & $[\times 10^4 s^{-1}]$ & $[s]$ & $(q_{10\%},q_{50\%},q_{90\%})$ & $[\times 10^4 s^{-1}]$\\
		\midrule
		\multirow{4}{*}{[-1.0, 1.0]}
		& RSM & 61.30 & (0.97, 1.00, 1.00) & 0.02 & 57.64 & (0.91, 1.00, 1.00) & 0.02  \\
		& ExpT & 2.30 & (0.98, 1.00, 1.00) & 0.43 & 2.83 & (0.96, 1.00, 1.00) & 0.34 \\
		& Gibbs & 19.70 & (0.84, 0.86, 0.91) & 0.04 & 21.18 & (0.75, 0.84, 0.91) & 0.04 \\
		& HMC & \textbf{1.89} & (0.95, 0.99, 1.00) & \textbf{0.50} & \textbf{1.92} & (0.94, 0.99, 1.00) & \textbf{0.49} \\
		\midrule
		\multirow{4}{*}{[-0.75, 0.75]}
		& RSM & 63.59 & (1.00, 1.00, 1.00) & 0.02 & 48.66 & (0.95, 0.99, 1.00) & 0.02  \\
		& ExpT & 3.22 & (0.96, 0.99, 1.00) & 0.30 & 3.24 & (0.98, 1.00, 1.00) & 0.30 \\
		& Gibbs & 20.20 & (0.83, 0.86, 0.91) & 0.04 & 18.23 & (0.74, 0.84, 0.93) & 0.04 \\
		& HMC & \textbf{1.46} & (0.94, 1.00, 1.00) & \textbf{0.64} & \textbf{1.28} & (0.94, 0.97, 1.00) & \textbf{0.73} \\
		\midrule
		\multirow{4}{*}{[-0.6, 0.6]}
		& RSM & 242.34 & (0.94, 0.97, 1.00) & 0 & 101.20 & (0.96, 1.00, 1.00) & 0.01  \\
		& ExpT & 2.94 & (0.94, 1.00, 1.00) & 0.32 & 2.80 & (0.98, 1.00, 1.00) & 0.35 \\
		& Gibbs & 18.89 & (0.80, 0.83, 0.94) & 0.04 & 18.90 & (0.77, 0.84, 0.92) & 0.04 \\
		& HMC & \textbf{1.72} & (0.92, 0.99, 1.00) & \textbf{0.53} & \textbf{1.68} & (0.93, 0.96, 1.00) & \textbf{0.55} \\
		\midrule
		\multirow{4}{*}{[-0.5, 0.5]}
		& RSM & 25512.77 & (0.98, 1.00, 1.00) & 0 & 157.06 & (0.96, 0.99, 1.00) & 0.01 \\
		& ExpT & \textbf{2.50} & (0.99, 1.00, 1.00) & \textbf{0.40} & \textbf{2.69} & (0.97, 1.00, 1.00) & \textbf{0.36} \\
		& Gibbs$^\dagger$ & --- & --- & --- & --- & --- & --- \\
		& HMC & 6.20 & (0.86, 0.90, 0.98) & 0.14 & 2.14 & (0.52, 0.85, 0.97) & 0.24 \\
		\bottomrule
	\end{tabular}
\end{table}

\begin{figure}[t!]
	\centering
	\includegraphics[width=0.4\textwidth]{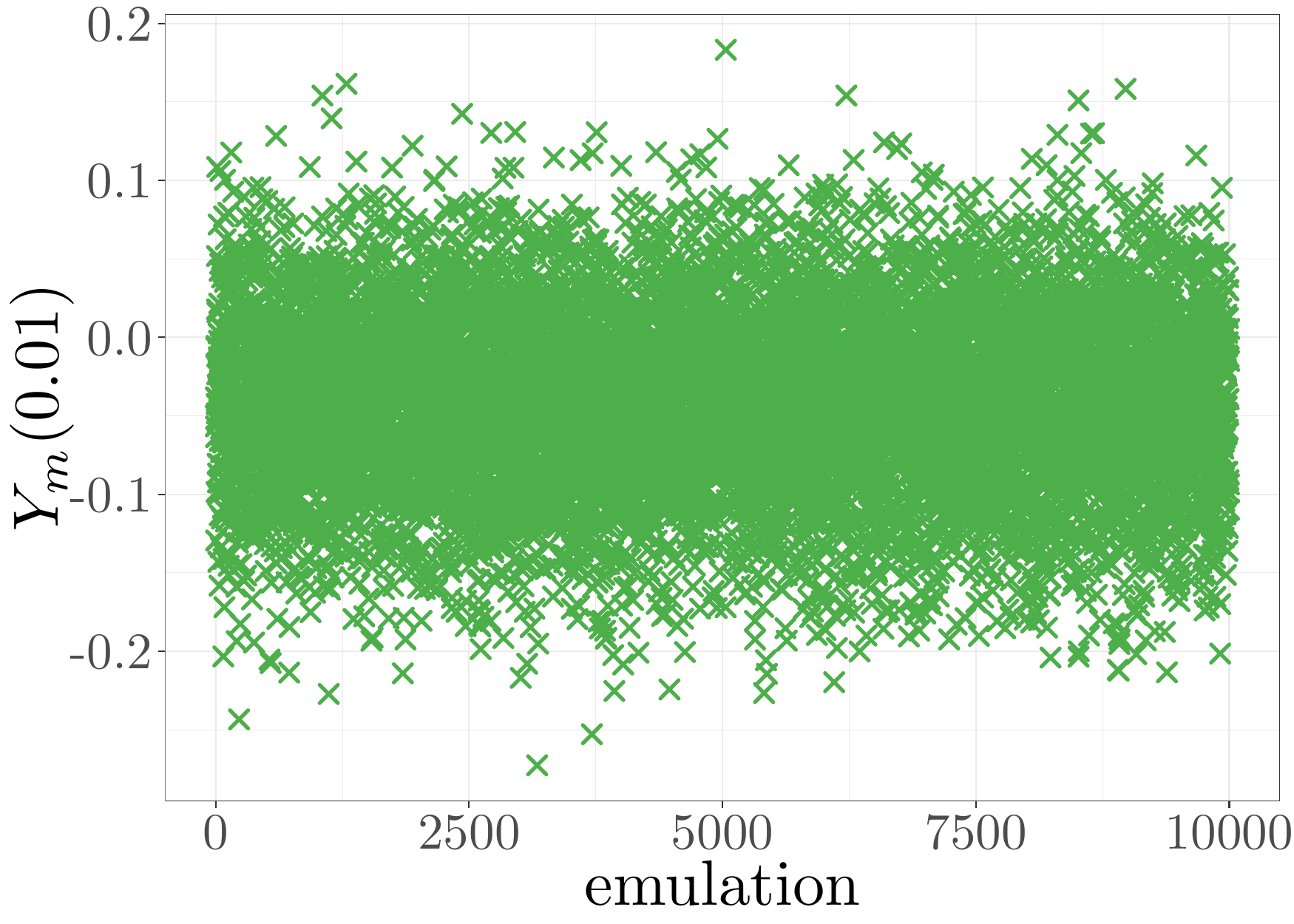}
	\includegraphics[width=0.4\textwidth]{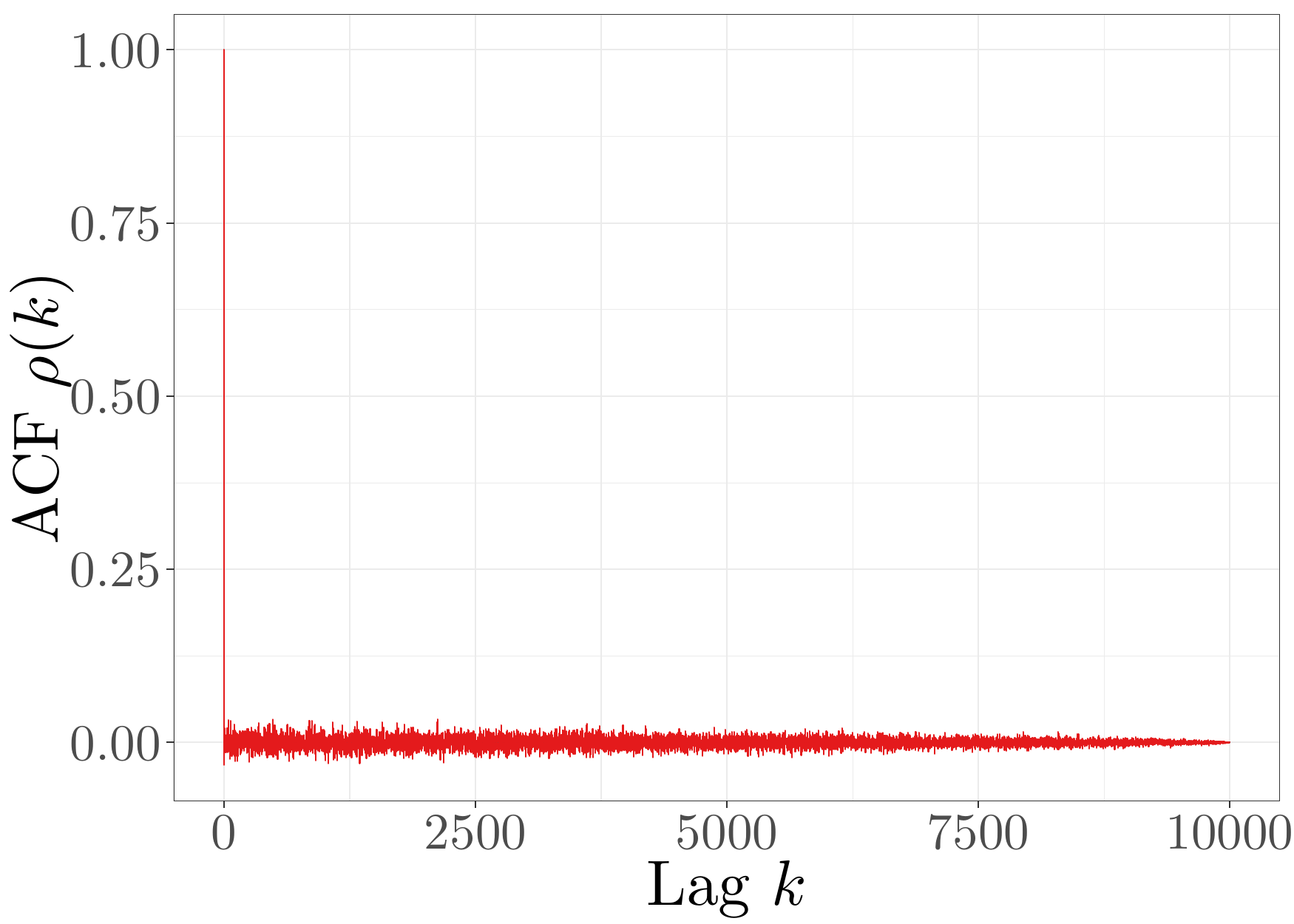}
	
	\caption{Efficiency of the HMC sampler in terms of its mixing performance. Results are shown for the  (left) trace and (right) autocorrelation plots at $Y_m(0.01)$.}
	\label{fig:MCMCmixing}
\end{figure}

Finally, we assess the efficiency of the HMC sampler in terms of its mixing performance (see Figure \ref{fig:MCMCmixing}). We analyse the example of Figure \ref{fig:Comparisson} using the noisy GP emulator with $\alpha = 0.5$. From both the trace and autocorrelation plots at $Y_m(0.01)$, one can conclude that the HMC sampler mixes well with small correlations.

\subsection{1D Toy Example under Multiple Constraints}
\label{subsec:toy1DMultiConstr}
In \citep{LopezLopera2017FiniteGPlinear}, numerical implementations were limited to noise-free observations that fulfilled the inequality constraints. In this example, we test the case when noisy observations do not necessarily satisfy the inequalities.

Consider the sigmoid function given by
\begin{equation}
x \mapsto \frac{1}{1 + \exp\big\{-10(x-\frac{1}{2})\big\}}, \ \mbox{ for } \ x \in [0, 1].
\label{eq:toyExample2}
\end{equation}
We evaluated~\eqref{eq:toyExample2} at $n = 300$ random values of $x$, and we contaminated the function evaluations with an additive Gaussian white noise with a standard deviation equal to 10\% of the sigmoid range. Since~\eqref{eq:toyExample2} exhibits both boundedness and non-decreasing conditions, we added those constraints into the GP emulator $Y_m$ using the convex set
\begin{equation*}
\mathcal{C}_{[0,1]}^{\uparrow} = \bigg\{\textbf{c} \in \realset{m}; \ \forall j = 2, \cdots, m \ : \  c_j \geq c_{j-1}, \ c_1 \geq 0, \ c_m \leq 1 \bigg\}.
\end{equation*}
Hence, the MC and MCMC samplers will be performed on $\realset{m+1}$ (number of inequality conditions). As a covariance function, we used a SE kernel, and we estimated the parameters $(\sigma^2, \ell, \tau^2)$ via ML.

\begin{figure}[t!]
	\centering
	\subfigure[\scriptsize\label{subfig:ToyExample2m5}$m=5$, $\text{CPU Time} = 0.03 \ 
	s$]{\includegraphics[width=0.325\textwidth]{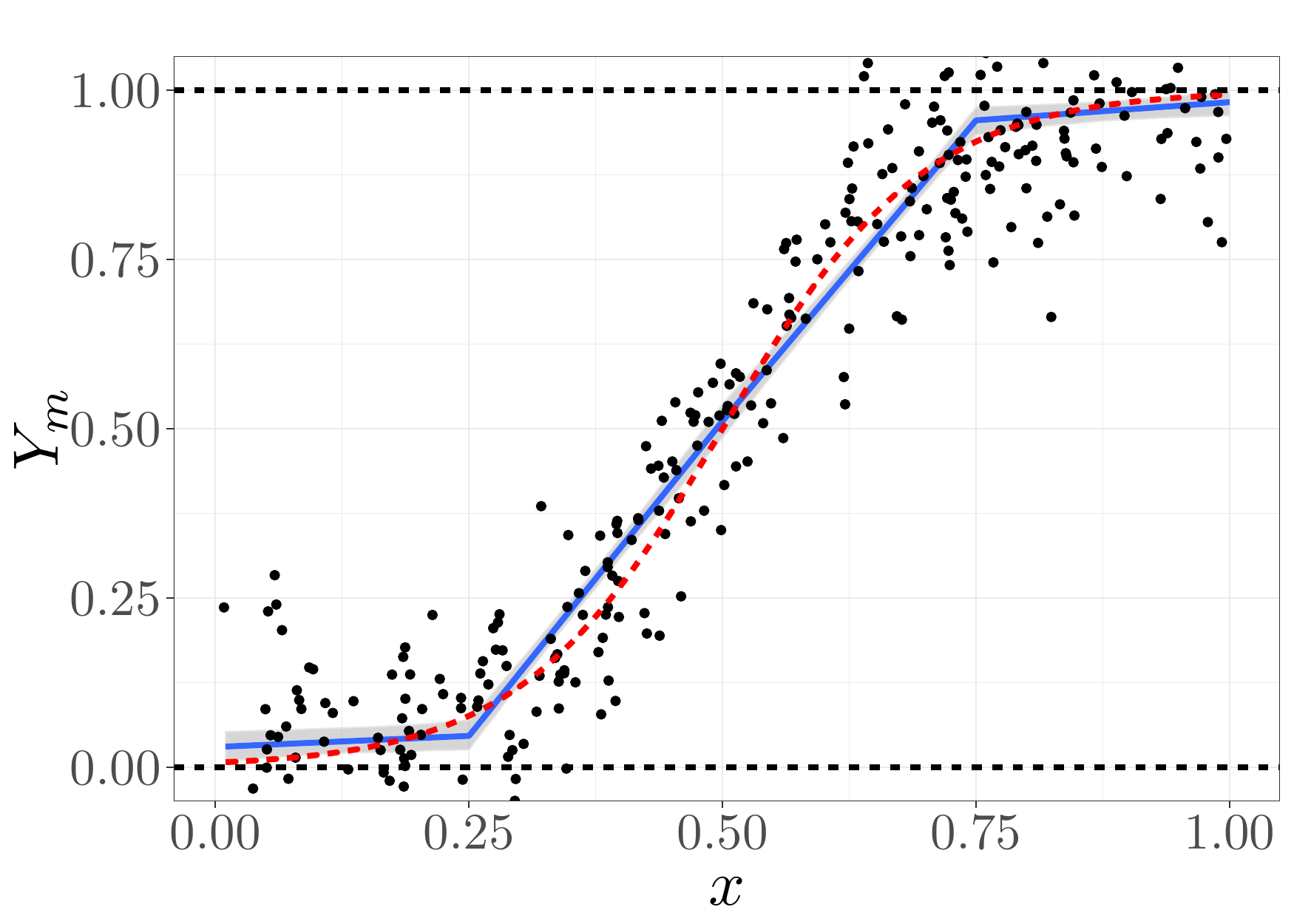}}
	\subfigure[\scriptsize\label{subfig:ToyExample2m25}$m=25$, $\text{CPU Time} = 0.09 \ s$]{\includegraphics[width=0.325\textwidth]{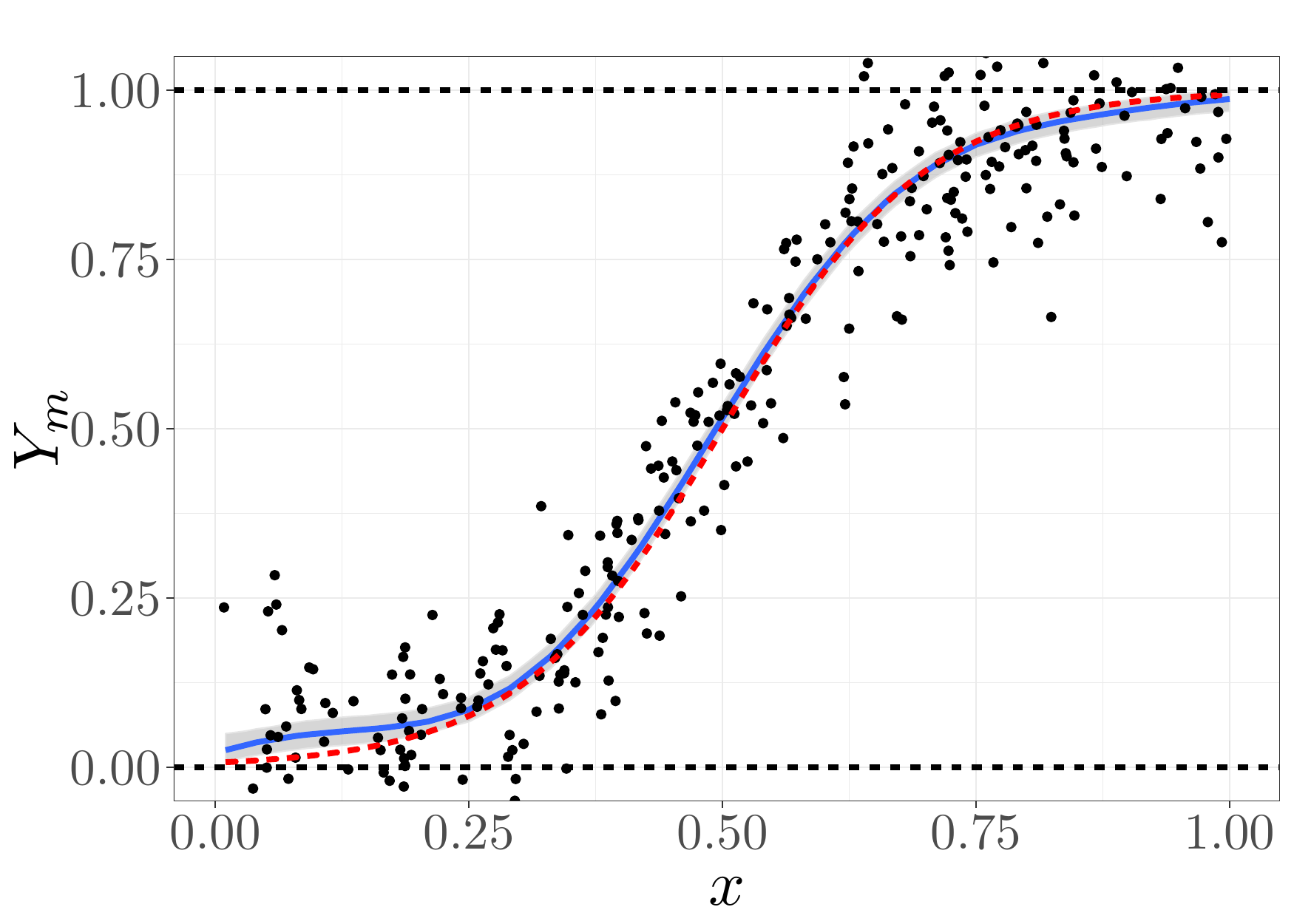}}
	%
	\subfigure[\scriptsize\label{subfig:ToyExample2m100}$m=100$, $\text{CPU Time} = 1.20 \ s$]{\includegraphics[width=0.325\textwidth]{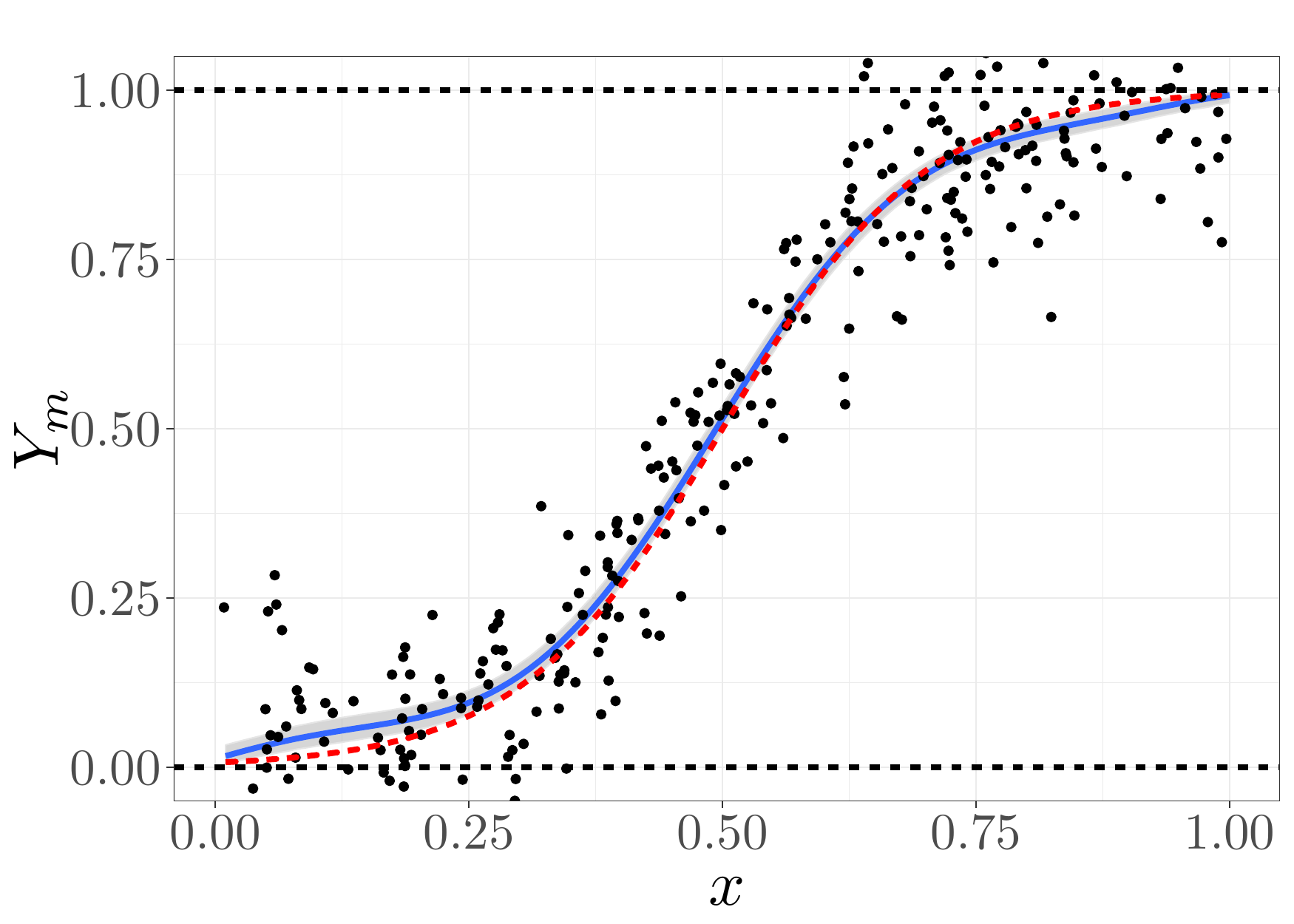}}
	
	\caption{GP emulators under boundedness and monotonicity constraints. Results are shown for different number values of knots $m$. Each panel shows: the target function (dashed lines), the noisy training points (dots), the conditional mean (solid line), the 95\% prediction interval (grey region), and the bounds (horizontal dashed lines).}
	\label{fig:ToyExample2}
\end{figure}

Unlike \citep{LopezLopera2017FiniteGPlinear}, there is no need here to satisfy the condition $m \geq n$, due to the noise. Therefore, the finite approximation of Section~\ref{sec:contrGPs} can be seen as a surrogate model of standard GP emulators for $m \ll n$. Figure~\ref{fig:ToyExample2} shows the performance of the constrained emulators via HMC for $m = 5, 25, 100$. For smaller values of $m$, the GP emulator runs fast but with a low quality of resolution of the approximation. For example, for $m = 5$, because of the linearity assumption between knots, the predictive mean presents breakpoints at the knots. On the other hand, the GP emulator yields smoother (constrained) emulations as $m$ increases ($m \geq 25$).  In particular, one can observe that for $m = 25$, the emulator leads to a good trade-off between quality of resolution and running time (13 times faster than for $m = 100$).

Finally, we test the performance of the proposed framework under different regularity assumptions, noise levels and inequality constraints. For the example in Figure \ref{fig:ToyExample2}, we fixed $m = 200$ and used different choices of covariance functions: either a Mat\'ern 3/2 kernel, a Mat\'ern 5/2 kernel or a SE kernel. Given a fixed noise level, the covariance parameters of each GP model, i.e. $\Btheta = (\sigma^2, \ell)$, were estimated via ML. The noise levels were chosen using different proportions of the sigmoid range. We assessed the proposed GP emulator accounting for either boundedness constraints, monotonicity constraints or both. We computed the CPU time and the $Q^2$ criterion. The $Q^2$ criterion is given by $Q^2 = 1 - \operatorname{SMSE}$, where SMSE is the standardised mean squared error \citep{Rasmussen2005GP}, and is equal to one if the predictive mean is equal to the test data and lower than one otherwise. We used the 300 noise-free function evaluations from \eqref{eq:toyExample2} as test data. Results are shown in Table \ref{tab:comparissonKernelNoise}. One can note that the introduction of noise let us also have constrained GP emulations in the cases where the regularity of the GP prior is not in agreement with the regularity of data and the inequality conditions. In particular, expensive procedures were obtained for the Mat\'ern 3/2 kernel when considering monotonicity. In those cases, the high irregularity of the (unconstrained) GP prior yielded more restrictive sample spaces that fulfil the monotonicity conditions. Furthermore, one may observe that the computational cost of emulators can be attenuated by increasing the noise level but at the cost of the accuracy of predictions.	

	\begin{table}[t!]
		\centering
		\footnotesize
		\caption{Performance of the GP emulators from Figure \ref{fig:ToyExample2} under different regularity assumptions, noise levels and inequality constraints. The noise levels were chosen using different proportions of the range of the sigmoid function in \eqref{eq:toyExample2}. CPU Time $[s]$ and $Q^2$ $[\%]$ results are shown for various covariance function (i.e. Mat\'ern 3/2 kernel, Mat\'ern 5/2 kernel and SE kernel), and different inequality constraints.}
		\label{tab:comparissonKernelNoise} \bigskip
		\begin{tabular}{c|cc|cc|cc}
			\toprule
			\multirow{2}{*}{Noise} & \multicolumn{6}{c}{Boundedness Constraints}  \\
			\multirow{2}{*}{level} & \multicolumn{2}{c|}{Mat\'ern$\frac{3}{2}$} & \multicolumn{2}{c|}{Mat\'ern$\frac{5}{2}$} & \multicolumn{2}{c}{SE} \\ \rule{0pt}{2.5ex}    
			& Time & $Q^2$ & Time & $Q^2$ & Time & $Q^2$ \\
			\midrule
			0\% 		& --- & ---  & --- & ---  & --- & --- \\	
			0.5\% 		& 1.0 & 99.4 & 0.8 & 99.6 & 0.6 & 99.7 \\
			1.0\% 		& 1.1 & 99.4 & 0.7 & 99.6 & 0.6 & 99.7 \\
			5.0\% 		& 1.0 & 98.9 & 0.8 & 99.3 & 0.6 & 99.5 \\
			10.0\% 	    & 0.9 & 98.2 & 0.8 & 98.9 & 0.6 & 99.2 \\
			\bottomrule
		\end{tabular} \hskip3ex
		\begin{tabular}{c|cc|cc|cc}
			\toprule
			\multirow{2}{*}{Noise} & \multicolumn{6}{c}{Monotonicity Constraints} \\
			\multirow{2}{*}{level} & \multicolumn{2}{c|}{Mat\'ern$\frac{3}{2}$} & \multicolumn{2}{c|}{Mat\'ern$\frac{5}{2}$} & \multicolumn{2}{c}{SE} \\
			& Time & $Q^2$ & Time & $Q^2$ & Time & $Q^2$ \\
			\midrule
			0\% 		& --- & ---  & --- & ---  & --- & --- \\	
			0.5\% 		& 117.0 & 99.5 & 1.4 & 99.8 & 1.2 & 99.8 \\
			1.0\% 		& 14.5  & 99.1 & 1.2 & 99.8 & 1.0 & 99.8 \\
			5.0\% 		&  7.4  & 95.6 & 1.0 & 99.3 & 0.8 & 99.3 \\
			10.0\% 	    &  6.3  & 91.9 & 1.0 & 98.7 & 0.6 & 98.9 \\
			\bottomrule
		\end{tabular}
		\bigskip
		
		\begin{tabular}{c|cc|cc|cc}
			\toprule
			\multirow{2}{*}{Noise} & \multicolumn{6}{c}{Boundedness \& Monotonicity Constraints} \\
			\multirow{2}{*}{level} & \multicolumn{2}{c|}{Mat\'ern$\frac{3}{2}$} & \multicolumn{2}{c|}{Mat\'ern$\frac{5}{2}$} & \multicolumn{2}{c}{SE} \\ \rule{0pt}{2.5ex}    
			& Time & $Q^2$ & Time & $Q^2$ & Time & $Q^2$ \\
			\midrule
			0\% 		& --- & ---  & --- & ---  & --- & --- \\	
			0.5\% 		& --- & --- & 17.3 & 99.7 & 13.9 & 99.8 \\
			1.0\% 		& $>10^4$ & 99.4 & 15.2 & 99.6 & 10.4 & 99.6 \\
			5.0\% 		& 251.8 & 96.7 & 13.3 & 98.6 & 8.6 & 98.3 \\
			10.0\% 	    & 246.1 & 94.6  &  13.3 & 97.5 & 8.6 & 97.0 \\
			\bottomrule
		\end{tabular}		
	\end{table}

\subsection{Coastal flooding applications}
\label{chap:lineqGPsNoisyObs:sec:app}



Coastal flooding models based on GP emulators have taken great attention regarding computational simplifications for estimating flooding indicators (like maximum water level at the coast, discharge, flood spatial extend, etc.) \citep{Rohmer2012CoastalFlooding,Azzimonti2018CoastalFlooding}. However, since standard GP emulators do not take into account the nature of many coastal flooding events satisfying positivity and/or monotonicity constraints, those approaches often require a large number of observations (commonly costly to obtain) in order to obtain reliable predictions. In those cases, GP emulators yield expensive procedures. Here we show that, by enforcing GP emulators to those inequality constraints, our framework can lead to more reliable prediction also when a small amount of data is available.

Here, we test the performance of the emulator in \eqref{eq:finApproxGPEmulator} on two coastal flooding datasets provided by the BRGM (which is the French Geological Survey, ``Bureau de Recherches G\'eologiques et Mini\`eres'' in French). The first dataset corresponds to a 2D coastal flooding application located on the Mediterranean coast, focusing on the water level at the coast \citep{Rohmer2012CoastalFlooding}. The second one describes a 5D coastal flooding example induced by overflow on the Atlantic coast, focusing on the inland flooded surface \citep{Azzimonti2018CoastalFlooding}. We trained different GP emulators whether the inequality constraints are considered or not. For the unconstrained emulators, we use the GP-based scheme provided by the R package \texttt{DiceKriging} \citep{Roustant2012DiceKriging}.
\begin{figure}[b!]
	\centering
	\includegraphics[width=0.35\textwidth]{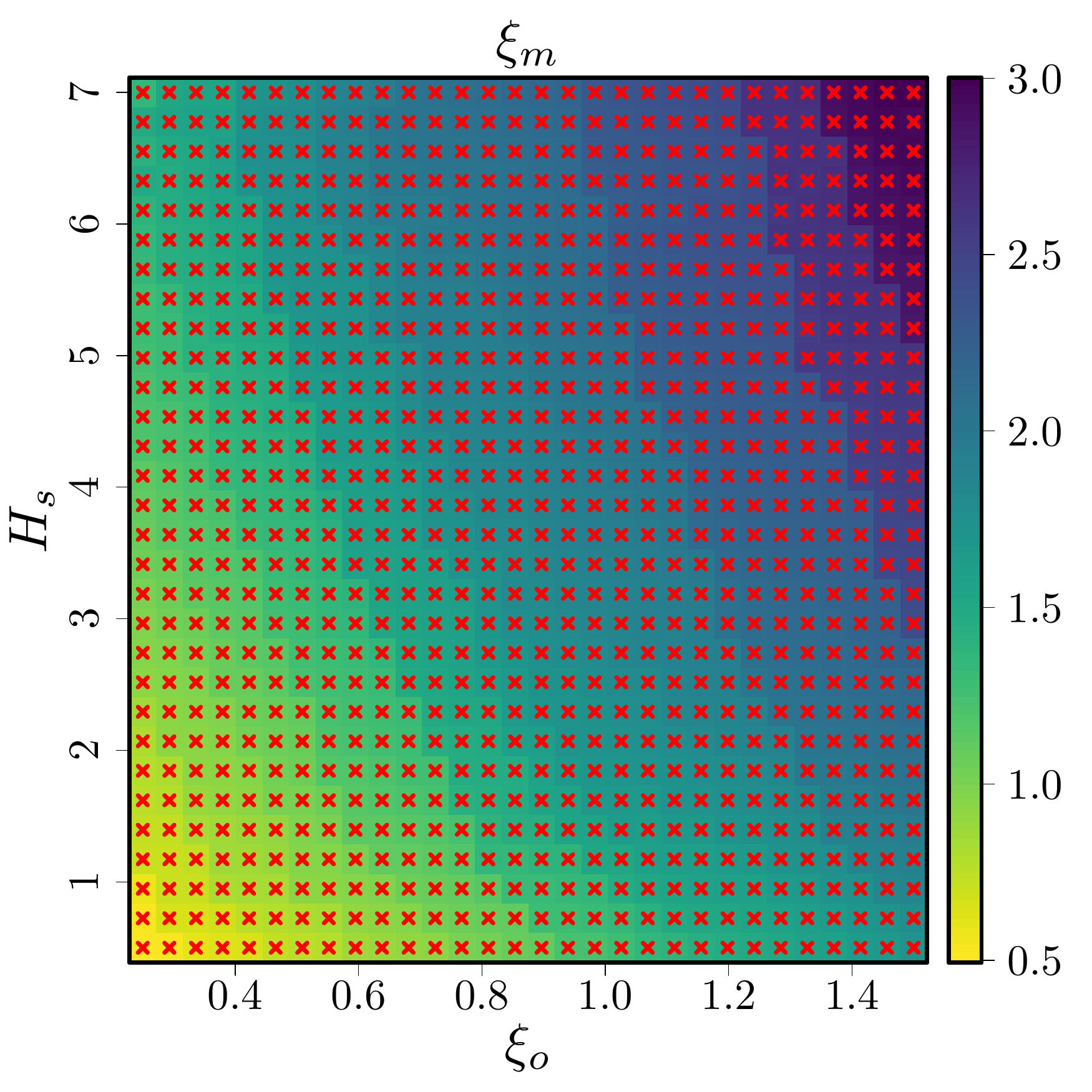} \hspace{20pt}
	\includegraphics[width=0.33\textwidth]{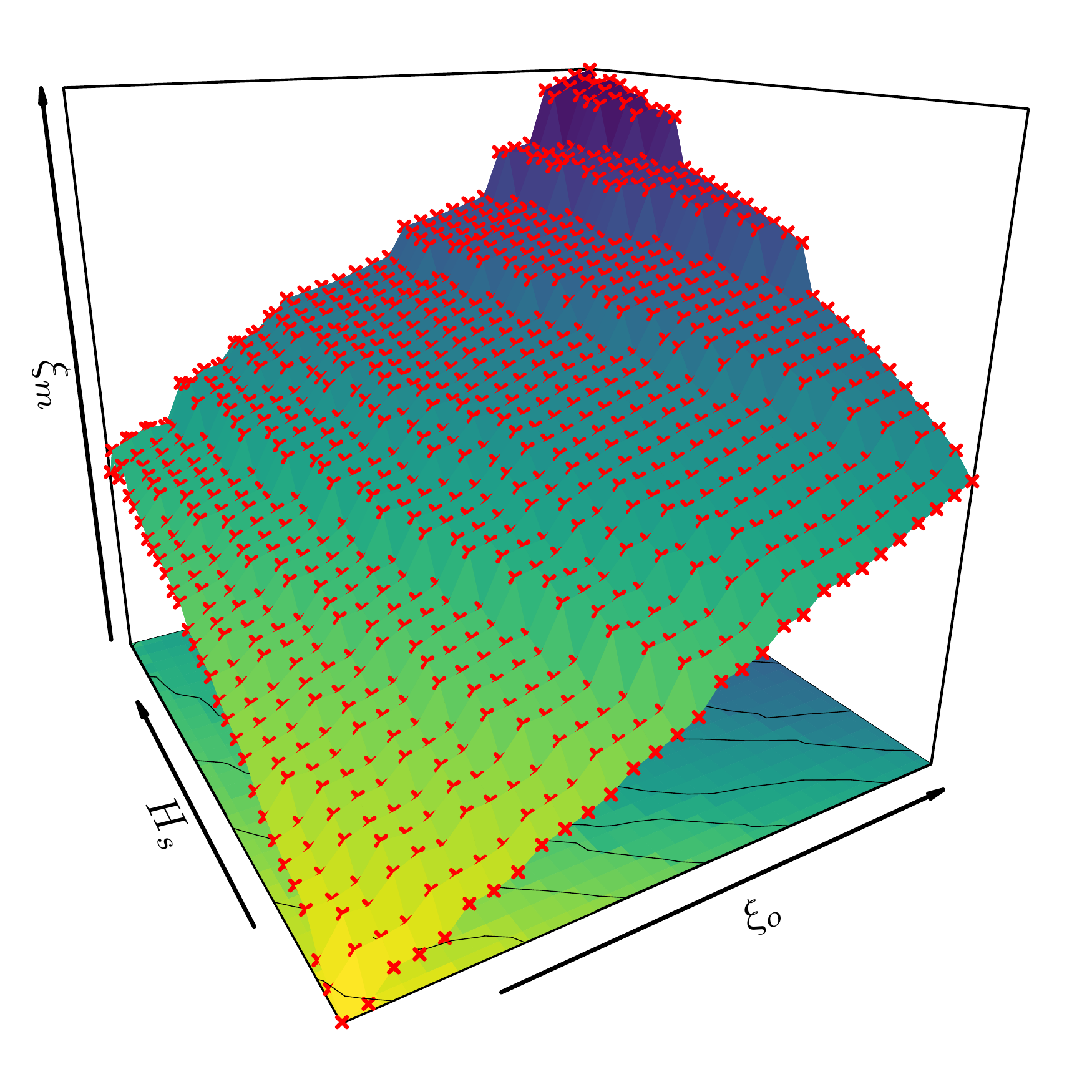}
	\caption[2D coastal flooding application]{2D coastal flooding application. (Left) 2D visualization of the $\xi_{m}$ values measured over a regular grid. (Right) 3D visualization of the $\xi_{m}$ data.}
	\label{fig:BRGM2DData}
\end{figure}

\subsubsection{2D application}
\label{subsec:lineqGPsNoisyObs:subsec:2Dapp}	
The coastal study site is located on a lido, which has faced two flood events in the past \citep{Rohmer2012CoastalFlooding}. The dataset used here contains 900 observations of the maximum water level at the coast $\xi_{m}$ depending on two input parameters: the offshore water level ($\xi_o$) and the wave height ($H_s$), both in metre units. The observations are taken within the domains $\xi_o \in [0.25, 1.50]$ and $H_s \in [0.5, 7]$ (with each dimension being discretized in 30 elements). One must note that, on the domain considered for the input variables, $\xi_{m}$ increases as $\xi_o$ and $H_s$ increase (see Figure \ref{fig:BRGM2DData}).

Here, we normalised the input space to be in $[0, 1]^2$. As covariance function, we used the tensor product of 1D SE kernels,
\begin{equation*}
k_{\Btheta}(\Bx,\Bx') = \sigma^2 \exp\Big\{- {\frac{(x_1-x'_1)^2}{2\ell_1^2}}\Big\} \exp\Big\{- {\frac{(x_2-x'_2)^2}{2\ell_2^2}}\Big\},
\end{equation*}
with covariance parameters $\Btheta = (\sigma^2, \ell_1, \ell_2)$. Both $\Btheta$ and the noise variance $\tau^2$ are estimated via ML. For the constrained model, we proposed emulators accounting for both positivity and monotonicity constraints, and we manually fixed the number of knots $m_1 = m_2 = 25$ aiming a trade-off between high quality of resolution and computational cost.
\begin{figure}[t!]
	\centering
	\begin{minipage}{0.41\textwidth}
		\centering
		\subfigure[\label{subfig:BRGM2DExampleSKMLEfig2} Unconstrained GP: $Q^2 = 0.987$]{\includegraphics[width=0.7\textwidth]{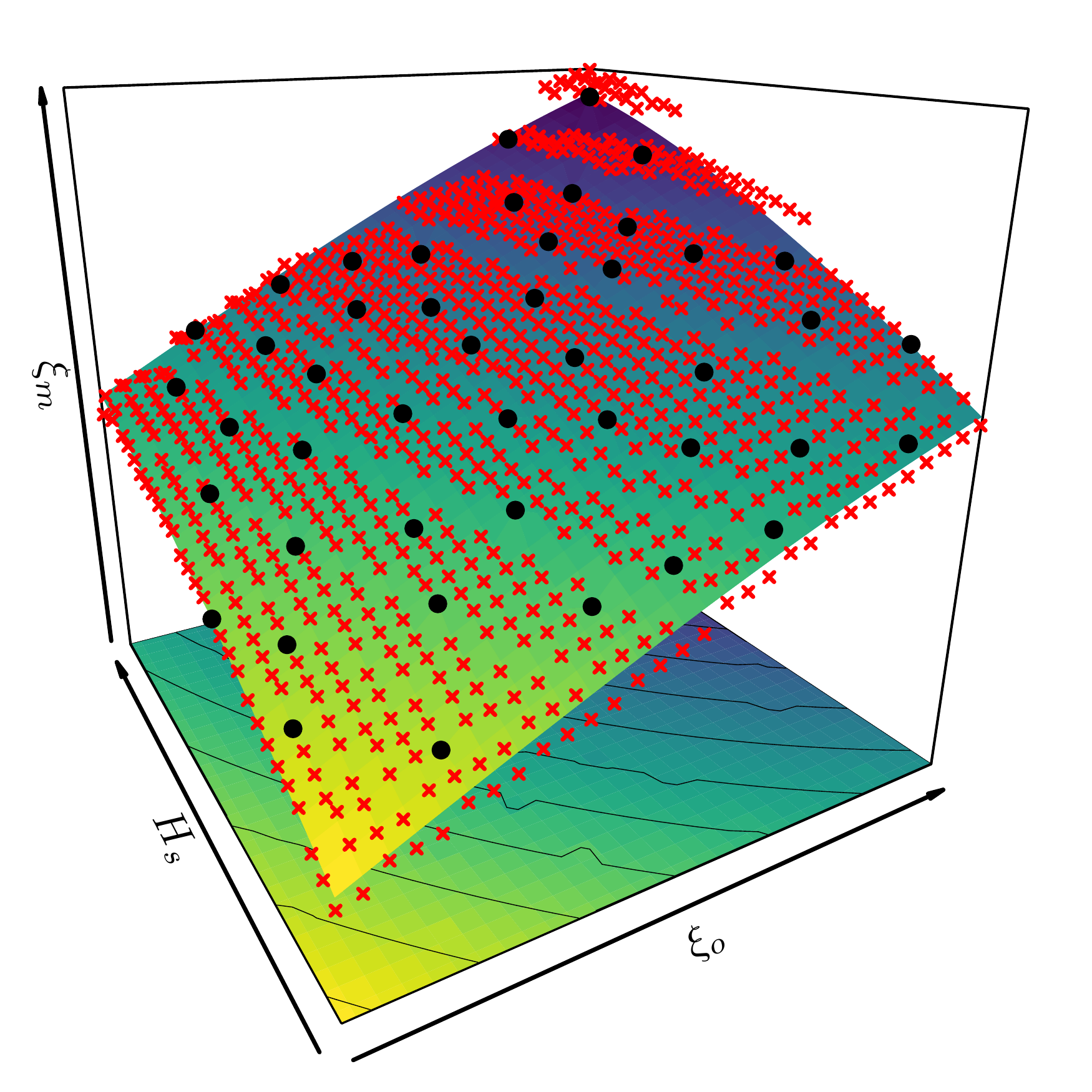}}
		
		\subfigure[\label{subfig:BRGM2DExampleCKMLEfig2} Constrained GP: $Q^2 = 0.991$]{\includegraphics[width=0.7\textwidth]{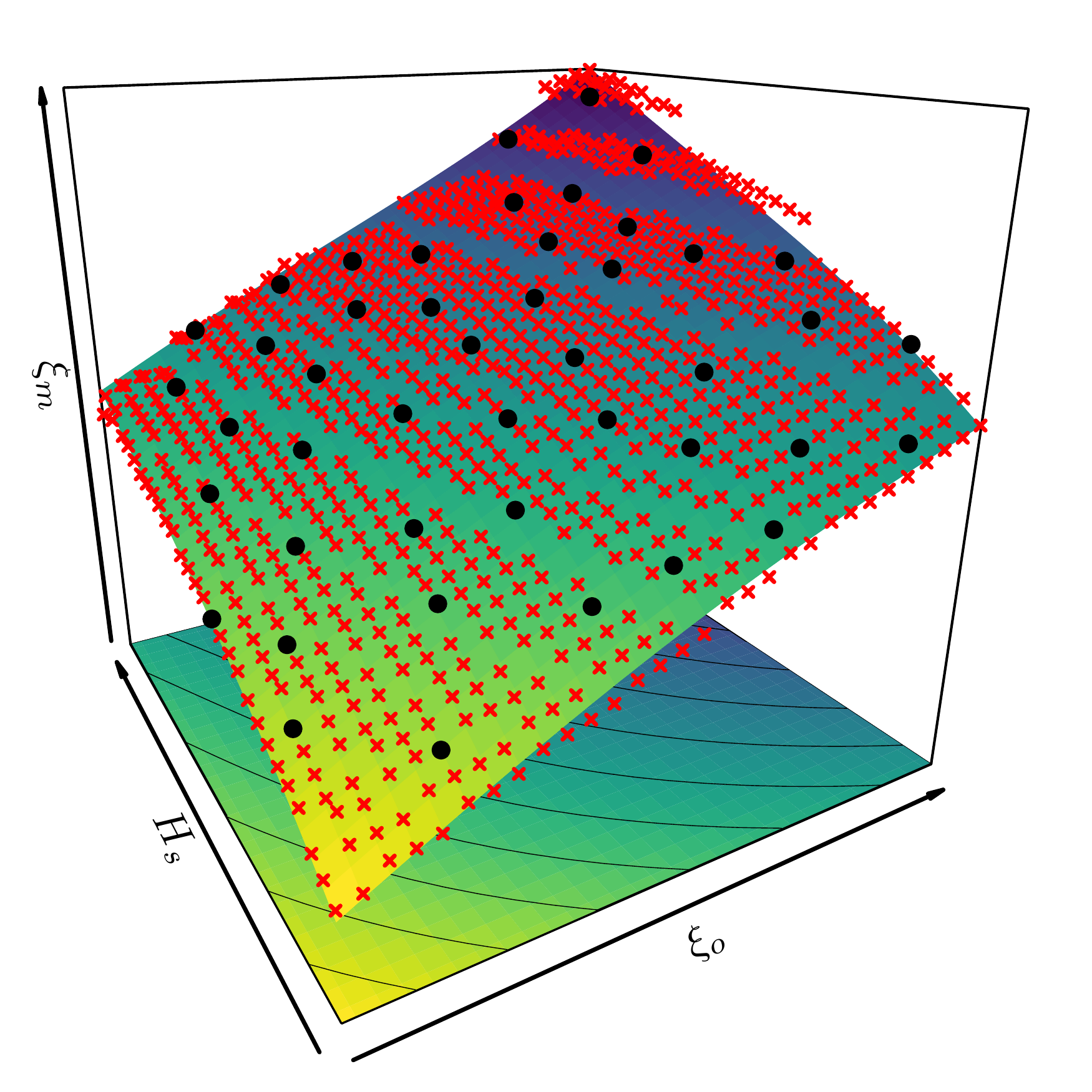}}
	\end{minipage}
	\begin{minipage}{0.58\textwidth}
		\centering
		\subfigure[\label{subfig:BRGM2DExampleSKMLEfig3} $Q^2$ performance]{\includegraphics[width=\textwidth]{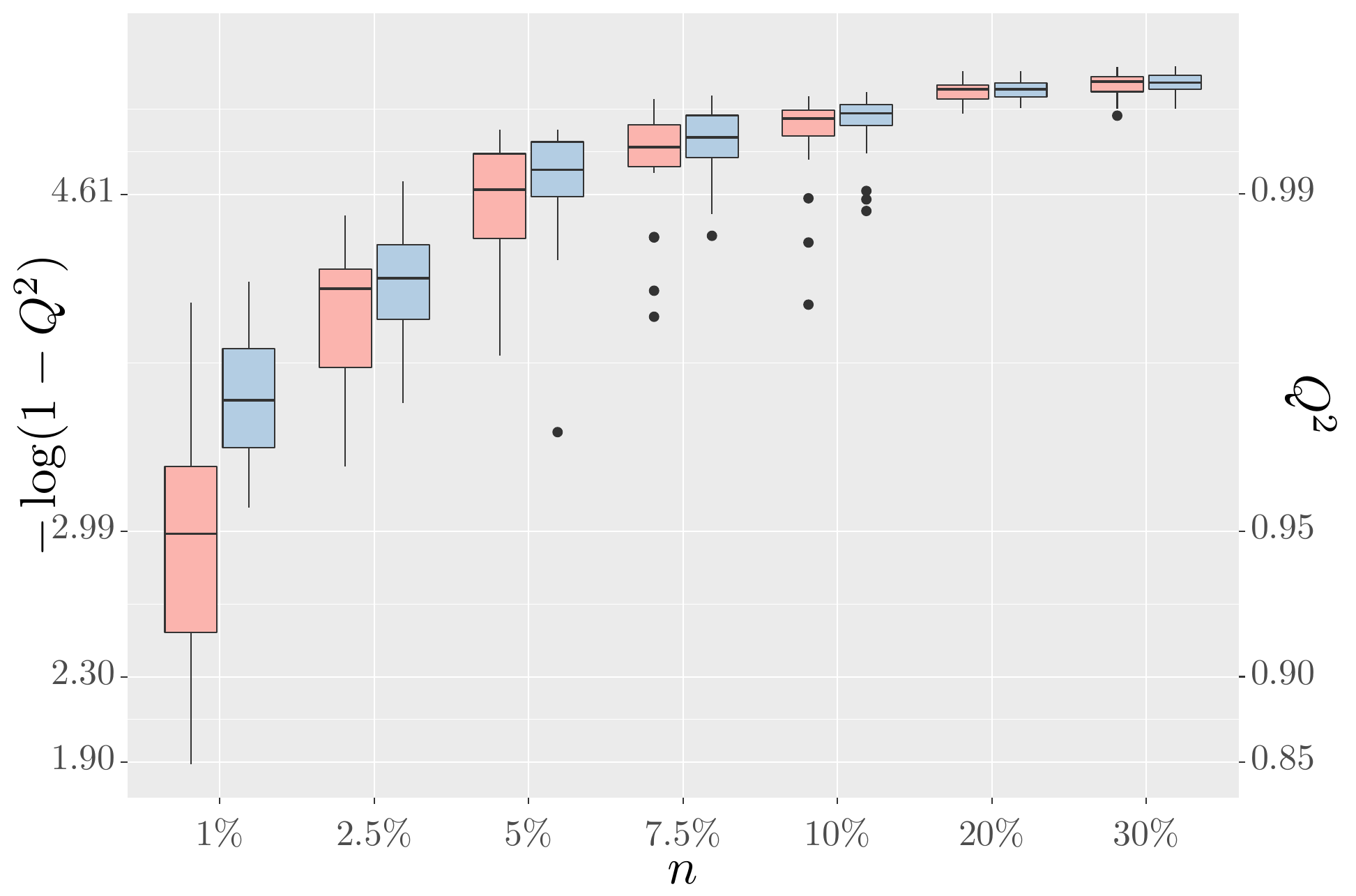}}
		\caption[2D GP emulators for modelling the coastal flooding data in \citep{Rohmer2012CoastalFlooding}]{2D GP emulators for modelling the coastal flooding data in \citep{Rohmer2012CoastalFlooding}. (Left) Prediction results using 5\% of the dataset via maximin Latin hypercube DoE. Each panel shows: training and test points (black dots and red crosses), the conditional mean function (solid surface), and the $Q^2$ criterion (subcaptions).  \subref{subfig:BRGM2DExampleSKMLEfig3} $Q^2$ assessment using different proportions of training points $n$ and using twenty different random training sets. Results are shown for the unconstrained (red) and constrained (blue) GP emulators.}
		\label{fig:BRGM2D}
	\end{minipage}			
\end{figure}	

For illustrative purposes, we first train both unconstrained and constrained GP emulators using 5\% of the data (equivalent to 45 training points chosen by a maximin Latin hypercube DoE), and we aim at predicting the remaining 95\%. Results are shown in Figures \ref{subfig:BRGM2DExampleSKMLEfig2} and \ref{subfig:BRGM2DExampleCKMLEfig2}. In particular, one can observe that the constrained GP emulator slightly outperformed the prediction around the extreme values of $\xi_m$, leading to an absolute improvement of $4\%$ of the $Q^2$ indicator. Then, we repeat the experiment using twenty different sets of training data and different proportions of training data. According to Figure \ref{subfig:BRGM2DExampleSKMLEfig3}, one can observe that the constrained emulator often outperforms the unconstrained one, with significant $Q^2$ improvements for small training sets. As coastal flooding simulators are commonly costly-to-evaluate, the benefit of having accurate prediction with lesser number of observations becomes useful for practical implementations. 

\subsubsection{5D application}
\label{chap:lineqGPsNoisyObs:subsec:5Dapp}

As in \citep{Azzimonti2018CoastalFlooding}, here we focus on the coastal flooding induced by overflow. We consider the ``Boucholeurs'' area located close to ``La Rochelle'', France. This area was flooded during the 2010 Xynthia storm, an event characterized by a high storm surge in phase with a high spring tide. We focus on those primary drivers, and on how they affect the resulting flooded surface. We refer to \citep{Azzimonti2018CoastalFlooding} for further details.

The dataset contains 200 observations of the flooded area $Y$ in $m^2$ depending on five input parameters $\textbf{x} = (T, S, \phi , t_{+} , t_{-})$ detailing the offshore forcing conditions:
\begin{itemize}
	\setlength\itemsep{0.2em}
	\item The tide is simplified by a sinusoidal signal parametrised by its high tide level $T \in [0.95, 3.70]$ ($m$).
	\item The surge signal is described by a triangular model using four parameters: the peak amplitude $S \in [0.65, 2.50]$ ($m$), the phase difference $\phi \in [-6, 6]$ (hours), between the surge peak and the high tide, the time duration of the raising part $t_{-} \in [-12.0, -0.5]$ (hours), and the falling part $t_{+} \in [0.5, 12.0]$ (hours).
\end{itemize}
The dataset is freely available in the R package \texttt{profExtrema} \citep{Azzimonti2018profExtrema}. One must note that the flooded area $Y$ increases as $T$ and $S$ increase.

Before implementing the corresponding GP emulators, we first analysed the structure of the dataset. We tested various standard linear regression models in order to understand the influence of each input variable $\textbf{x} = (T, S, \phi , t_{+} , t_{-})$. We assessed the quality of the linear models using the adjusted $R^2$ criterion. Similarly to the $Q^2$ criterion, the $R^2$ indicator evaluates the quality of predictions over all the observation points rather than only over the training data. Therefore, for noise-free observations, the $R^2$ indicator is equal to one if the predictors are exactly equal to the data. We also tested various models considering different input variables (e.g. transformation of variables, or inclusion of interaction terms). After testing different linear models, we observed that they were more sensitive to the inputs $T$ and $S$ rather than to other ones. We also noted that, by transforming the phase coordinate $\phi \mapsto \cos(2\pi\phi)$, an absolute improvement about 26\% of the $R^2$ indicator was obtained, and the influence of both $t_{-}$ and $t_{+}$ becomes more significant. Finally, we used these settings for the GP implementations.

We normalised the input space to be in $[0, 1]^5$, and we used a covariance function given by the Kronecker product of 1D Mat\'ern 5/2 kernels. The covariance parameters $\Btheta = (\sigma^2, \ell_1, \cdots, \ell_5)$ and the noise variance $\tau^2$ were estimated via ML. We also tested other types of kernel structures, including SE and Mat\'ern 3/2 kernels, but less accurate predictions were obtained according to the $Q^2$ criterion. For the constrained model, we proposed GP emulators accounting for positivity constraints everywhere. We also imposed monotonicity constraints along the $T$ and $S$ input dimensions. Since the computational complexity of the constrained GP emulator increases with the number of knots $m$ used in the piecewise-linear representation, we strategically fixed them in coordinates requiring high quality of resolution. Since we observed that the contribution of the inputs $T$, $S$, $t_{-}$ and $t_{+}$ was almost linear \citep[result in agreement with][]{Azzimonti2018CoastalFlooding}, we placed fewer number of knots over those entries. In particular, we fixed as number of knots per dimension: $m_1 = m_2 = 4$, $m_3 = 5$ and $m_1 = m_2 = 3$.

\begin{figure}[t!]
	\centering
	\begin{minipage}{0.59\textwidth}
		\centering		
		{\includegraphics[width=0.8\textwidth]{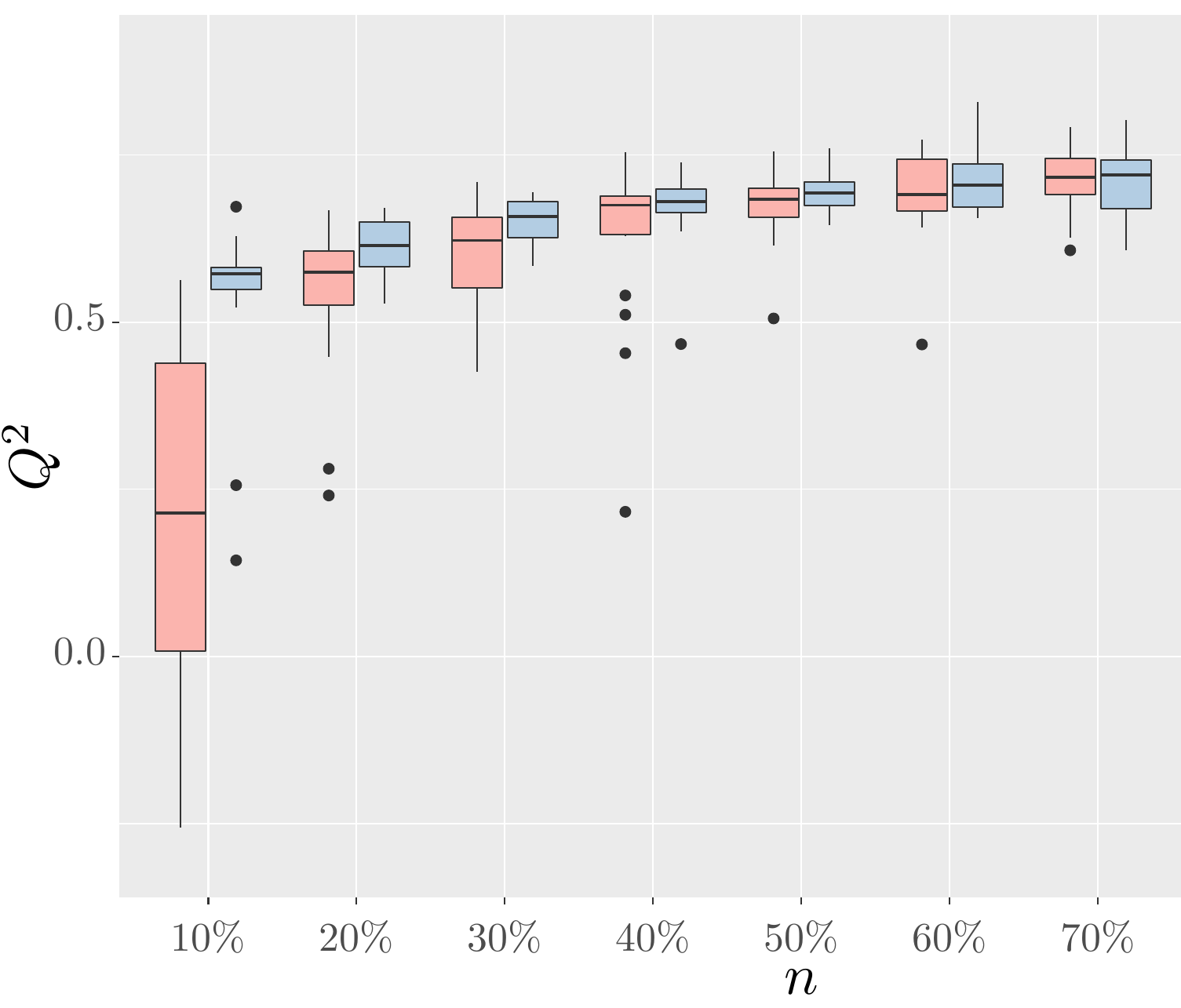}}
	\end{minipage}
	\hskip2ex
	\begin{minipage}{0.38\textwidth}
		\vskip-3ex
		\caption[5D GP emulators for modelling the coastal flooding data in \citep{Azzimonti2018CoastalFlooding}]{5D GP emulators for modelling the coastal flooding data in \citep{Azzimonti2018CoastalFlooding}. The boxplots show the $Q^2$ results using different proportions of training points $n$ and using twenty different random training sets. Results are shown for the unconstrained (red) and constrained (blue) GP emulators.}
		\label{fig:BRGM5D}
	\end{minipage}			
\end{figure}

As in Section \ref{subsec:lineqGPsNoisyObs:subsec:2Dapp}, we trained GP emulators using twenty different sets of training data and different proportions of training data. According to Figure \ref{fig:BRGM5D}, one can observe once again that the constrained GP emulator often outperforms the unconstrained one, with significant $Q^2$ improvements for small training sets. In particular, one can note that, by enforcing the GP emulators with both positivity and monotonicity constraints, accurate predictions were also provided by using only 10\% of the observations as training points (equivalent to 20 observations). 

\section{Conclusions}
\label{sec:conclusions}
We have introduced a constrained GP emulator with linear inequality conditions and noisy observations. By relaxing the interpolation of observations through a noise effect, MC and MCMC samplers are performed in less restrictive sample spaces. This leads to faster emulators while preserving high effective sampling rates. As seen in the experiments, the Hamiltonian Monte Carlo sampler from \citep{Pakman2014Hamiltonian} usually outperformed its competitors, providing much more efficient effective sample rates in high dimensional sample spaces. 

Since there is no need of having more knots than observations ($m \geq n$), the computational complexity of MC and MCMC samplers is independent of $n$. Therefore, since the samplers are performed on $\realset{m}$, they can be used for large values of $n$ by letting $m \ll n$. As shown in the 5D monotonic example, effective monotone emulations can be obtained within reasonable running times (about tens of minutes).

Despite the improvements obtained here for scaling the monotonic GP emulator in higher dimensions, its tensor structure makes it impractical for tens of input variables. We believe that this limitation could be mitigated by using other types of designs of the knots (e.g. sparse designs). In addition, supplementary assumptions on the nature of the target function can also be made to reduce the dimensionality of the sample spaces where MC and MCMC samplers are performed (e.g. additivity).

\section*{Acknowledgement}
This research was conducted within the frame of the Chair in Applied Mathematics OQUAIDO, gathering partners in technological research (BRGM, CEA, IFPEN, IRSN, Safran, Storengy) and academia (CNRS, Ecole Centrale de Lyon, Mines Saint-Etienne, University of Grenoble, University of Nice, University of Toulouse) around advanced methods for Computer Experiments.

\bibliographystyle{apa}  
\bibliography{arXiv2019_lineqGPNoise}  

\begin{thebibliography}{}

\bibitem[\protect\astroncite{Azzimonti}{2018}]{Azzimonti2018profExtrema}
Azzimonti, D. (2018).
\newblock prof{E}xtrema: Compute and visualize profile extrema functions.
\newblock \url{https://cran.r-project.org/web/packages/profExtrema/index.html}.

\bibitem[\protect\astroncite{Azzimonti
  et~al.}{2019}]{Azzimonti2018CoastalFlooding}
Azzimonti, D., Ginsbourger, D., Rohmer, J., and Idier, D. (2019).
\newblock Profile extrema for visualizing and quantifying uncertainties on
  excursion regions. {A}pplication to coastal flooding.
\newblock {\em Technometrics}, 0(ja):1--26.

\bibitem[\protect\astroncite{Bay et~al.}{2016}]{Bay2016KimeldorfWahba}
Bay, X., Grammont, L., and Maatouk, H. (2016).
\newblock {Generalization of the Kimeldorf-Wahba correspondence for constrained
  interpolation}.
\newblock {\em {Electronic journal of statistics }}, 10(1):1580--1595.

\bibitem[\protect\astroncite{Bishop}{2007}]{Bishop2007ML}
Bishop, C.~M. (2007).
\newblock {\em Pattern Recognition And Machine Learning (Information Science
  And Statistics)}.
\newblock Springer.

\bibitem[\protect\astroncite{Botev}{2017}]{Botev2017MinimaxTilting}
Botev, Z.~I. (2017).
\newblock The normal law under linear restrictions: simulation and estimation
  via minimax tilting.
\newblock {\em Journal of the Royal Statistical Society: Series B (Statistical
  Methodology)}, 79(1):125--148.

\bibitem[\protect\astroncite{Cousin et~al.}{2016}]{Cousin2016KrigingFinancial}
Cousin, A., Maatouk, H., and Rulli\`ere, D. (2016).
\newblock Kriging of financial term-structures.
\newblock {\em European Journal of Operational Research}, 255(2):631--648.

\bibitem[\protect\astroncite{Deville et~al.}{2015}]{Deville2015kergp}
Deville, Y., Ginsbourger, D., Roustant, O., and Durrande, N. (2015).
\newblock {kergp: Gaussian Process Models with Customised Covariance Kernels.}
\newblock \url{https://cran.r-project.org/web/packages/kergp/index.html}.

\bibitem[\protect\astroncite{Dupuy et~al.}{2015}]{Dupuy2015DiceDesign}
Dupuy, D., Helbert, C., and Franco, J. (2015).
\newblock {DiceDesign} and {DiceEval}: Two {R} packages for design and analysis
  of computer experiments.
\newblock {\em Journal of Statistical Software}, 65(i11).

\bibitem[\protect\astroncite{Geyer}{1992}]{Geyer1992MCMC}
Geyer, C.~J. (1992).
\newblock {Practical Markov Chain Monte Carlo}.
\newblock {\em Statistical Science}, 7(4):473--483.

\bibitem[\protect\astroncite{Golchi et~al.}{2015}]{Golchi2015MonotoneEmulation}
Golchi, S., Bingham, D.~R., Chipman, H., and Campbell, D.~A. (2015).
\newblock Monotone emulation of computer experiments.
\newblock {\em SIAM/ASA Journal on Uncertainty Quantification}, 3(1):370--392.

\bibitem[\protect\astroncite{Goldfarb and Idnani}{1982}]{Goldfarb1982QP}
Goldfarb, D. and Idnani, A. (1982).
\newblock Dual and primal-dual methods for solving strictly convex quadratic
  programs.
\newblock In Hennart, J.~P., editor, {\em Numerical Analysis}, pages 226--239,
  Berlin, Heidelberg. Springer Berlin Heidelberg.

\bibitem[\protect\astroncite{Gong and Flegal}{2016}]{Gong2016ESS}
Gong, L. and Flegal, J.~M. (2016).
\newblock {A practical sequential stopping rule for high-dimensional Markov
  Chain Monte Carlo}.
\newblock {\em Journal of Computational and Graphical Statistics},
  25(3):684--700.

\bibitem[\protect\astroncite{Lan and Shahbaba}{2016}]{Lan2016Sampling}
Lan, S. and Shahbaba, B. (2016).
\newblock {\em Sampling Constrained Probability Distributions Using Spherical
  Augmentation}, pages 25--71.
\newblock Springer International Publishing, Cham.

\bibitem[\protect\astroncite{Larson and Bengzon}{2013}]{Larson2013FEM}
Larson, M.~G. and Bengzon, F. (2013).
\newblock {\em The Finite Element Method: Theory, Implementation, and
  Applications}.
\newblock Springer Publishing Company, Incorporated.

\bibitem[\protect\astroncite{L\'opez-Lopera}{2018}]{LopezLopera2018LineqGPR}
L\'opez-Lopera, A.~F. (2018).
\newblock lineq{GPR}: Gaussian process regression models with linear inequality
  constraints.
\newblock \url{https://cran.r-project.org/web/packages/lineqGPR/index.html}.

\bibitem[\protect\astroncite{L\'opez-Lopera
  et~al.}{2018}]{LopezLopera2017FiniteGPlinear}
L\'opez-Lopera, A.~F., Bachoc, F., Durrande, N., and Roustant, O. (2018).
\newblock Finite-dimensional {G}aussian approximation with linear inequality
  constraints.
\newblock {\em SIAM/ASA Journal on Uncertainty Quantification},
  6(3):1224--1255.

\bibitem[\protect\astroncite{Maatouk and Bay}{2016}]{Maatouk2016RSM}
Maatouk, H. and Bay, X. (2016).
\newblock {\em A New Rejection Sampling Method for Truncated Multivariate
  Gaussian Random Variables Restricted to Convex Sets}, pages 521--530.
\newblock Springer International Publishing, Cham.

\bibitem[\protect\astroncite{Maatouk and Bay}{2017}]{Maatouk2017GPineqconst}
Maatouk, H. and Bay, X. (2017).
\newblock Gaussian process emulators for computer experiments with inequality
  constraints.
\newblock {\em Mathematical Geosciences}, 49(5):557--582.

\bibitem[\protect\astroncite{Murphy}{2012}]{Murphy2012ML}
Murphy, K.~P. (2012).
\newblock {\em Machine Learning: A Probabilistic Perspective (Adaptive
  Computation And Machine Learning Series)}.
\newblock The MIT Press.

\bibitem[\protect\astroncite{Pakman and Paninski}{2014}]{Pakman2014Hamiltonian}
Pakman, A. and Paninski, L. (2014).
\newblock Exact {Hamiltonian Monte Carlo} for truncated multivariate
  {G}aussians.
\newblock {\em Journal of Computational and Graphical Statistics},
  23(2):518--542.

\bibitem[\protect\astroncite{Press et~al.}{1992}]{Press1992Num}
Press, W.~H., Teukolsky, S.~A., Vetterling, W.~T., and Flannery, B.~P. (1992).
\newblock {\em Numerical Recipes in C (2Nd Ed.): The Art of Scientific
  Computing}.
\newblock Cambridge University Press, New York, NY, USA.

\bibitem[\protect\astroncite{Rasmussen and Williams}{2005}]{Rasmussen2005GP}
Rasmussen, C.~E. and Williams, C. K.~I. (2005).
\newblock {\em Gaussian Processes for Machine Learning (Adaptive Computation
  and Machine Learning)}.
\newblock The MIT Press.

\bibitem[\protect\astroncite{Rohmer and
  Idier}{2012}]{Rohmer2012CoastalFlooding}
Rohmer, J. and Idier, D. (2012).
\newblock A meta-modelling strategy to identify the critical offshore
  conditions for coastal flooding.
\newblock {\em Natural Hazards and Earth System Sciences}, 12(9):2943--2955.

\bibitem[\protect\astroncite{Roustant et~al.}{2012}]{Roustant2012DiceKriging}
Roustant, O., Ginsbourger, D., Deville, Y., et~al. (2012).
\newblock {DiceKriging}, {DiceOptim}: Two {R} packages for the analysis of
  computer experiments by {Kriging}-based metamodeling and optimization.
\newblock {\em Journal of Statistical Software}, 51(i01).

\bibitem[\protect\astroncite{Taylor and
  Benjamini}{2017}]{Benjamini2017fastGibbs}
Taylor, J. and Benjamini, Y. (2017).
\newblock Restricted{MVN}: multivariate normal restricted by affine
  constraints.
\newblock
  \url{https://cran.r-project.org/web/packages/restrictedMVN/index.html}.
\newblock [Online; 02-Feb-2017].

\end{thebibliography}

\end{document}